\documentclass{article}
\usepackage{amssymb}

\usepackage[final]{corl_2025} % Uncomment for pre-prints (e.g., arxiv); This is like ``final'', but will remove the CORL footnote.

\usepackage{amsmath,amsfonts,bm}

\def\eqref#1{equation~\ref{#1}}
\def\1{\bm{1}}

\newcommand{\test}{\mathcal{D_{\mathrm{test}}}}

\DeclareMathAlphabet{\mathsfit}{\encodingdefault}{\sfdefault}{m}{sl}
\SetMathAlphabet{\mathsfit}{bold}{\encodingdefault}{\sfdefault}{bx}{n}

\def\gA{{\mathcal{A}}}

\def\gC{{\mathcal{C}}}
\def\gD{{\mathcal{D}}}

\def\gG{{\mathcal{G}}}

\def\gM{{\mathcal{M}}}

\def\gO{{\mathcal{O}}}
\def\gP{{\mathcal{P}}}

\def\gT{{\mathcal{T}}}

\def\gV{{\mathcal{V}}}

\def\gX{{\mathcal{X}}}

\DeclareMathOperator*{\argmax}{arg\,max}

\usepackage{wrapfig}
\usepackage[linesnumbered,ruled,vlined] {algorithm2e}
\usepackage{graphicx}

\usepackage{fancyvrb}
\usepackage{fvextra}

\usepackage[shortlabels]{enumitem}
\usepackage[most]{tcolorbox}

\newtcolorbox{promptbox}{
  enhanced,
  breakable,             % allows page breaking
  colback=gray!10,       % background color
  colframe=gray!50,      % border color
  arc=2mm,               % rounded corners
  boxrule=0.4pt,         % border thickness
  verbatim,              % this enables true \begin{verbatim} behavior
  sharp corners=all,     % optional: control corner style
  fontupper=\ttfamily\small % monospace font
}

\newtheorem{definition}{Definition}

\title{From Real World to Logic and Back: \\ Learning Generalizable Relational Concepts For Long Horizon Robot Planning}

\author{
    Naman Shah$^{1,2}$, Jayesh Nagpal$^1$, and Siddharth Srivastava$^1$ \\ 
    $^1$School of Computing and Augmented Intelligence,
    Arizona State University, USA\\
    $^2$Department of Computer Science, Brown University, USA
}

\begin{document}
\maketitle

%===============================================================================

\begin{abstract}
	% Humans excel at generalizing and transferring knowledge from naïve demonstrations to complex tasks. However,  this still remains challenging for robots  fail to achieve human-level generalization due to existing data-inefficient behavior cloning approaches that rely solely on demonstrations. This paper presents the first known approach for learning symbolic generative concepts from a small number of simple unlabeled and unsegmented training demonstrations, enabling generalization to much more complex tasks. Through empirical results in both real and simulated settings, we show that the autonomously generated "world models" allow robots to solve novel, unseen test tasks with significantly longer execution horizons and many more objects in zero-shot fashion.

Robots still lag behind humans in their ability to generalize from limited experience, particularly when transferring learned behaviors to long-horizon tasks in unseen environments. We present the first method that enables robots to autonomously invent symbolic, relational concepts directly from a small number of raw, unsegmented, and unannotated demonstrations. From these, the robot learns logic-based world models that support zero-shot generalization to tasks of far greater complexity than those in training. Our framework achieves performance on par with hand-engineered symbolic models, while scaling to execution horizons far beyond training and handling up to 18$\times$ more objects than seen during learning. The results demonstrate a framework for autonomously acquiring transferable symbolic abstractions from raw robot experience, contributing toward the development of interpretable, scalable, and generalizable robot planning systems. Project website and code: \url{https://aair-lab.github.io/r2l-lamp}.

% Humans are able to learn to solve complex tasks based on simple demonstrations. However, this is a challenging problem for robotics; current training data requirements make it difficult to develop reliable autonomous robots for settings where test tasks feature more objects and longer horizons than those encountered during learning. We show that enabling a robot to autonomously invent well-founded concepts can allow it to learn and transfer knowledge more efficiently from small batches of demonstrations on simple tasks. The robot is then able to learn world models and solve new, complex tasks at a level comparable with hand-crafted models. Empirical results indicate that this framework allows robots to solve new tasks with much longer
% execution horizons and many more objects (up to 18x) than those encountered in
% demonstrations, in zero-shot fashion. This yields the first known approach for autonomously inventing relational concepts and for learning logic-based generalizable world models from small batches of unannotated high-dimensional, real-valued robot demonstrations.
\end{abstract}

% Two or three meaningful keywords should be added here

%===============================================================================

\section{Introduction}
	
The ability to learn from simple
examples or demonstrations, and to generalize and transfer that knowledge to solve more complex problems
(a hallmark of learning in humans) is a challenging robot learning problem. 
% Learning how to accomplish complex tasks is essential for robots to safely
% assist humans in a wide variety of domains. 
 For instance, all the concepts necessary
for clearing a table cluttered with cups are present in one robot trajectory
that picks up and places a cup (with no annotations or segmentation). It should
therefore be possible to clear a table scattered with cups, given a handful of
demonstrations for picking and placing a single cup. 
Yet, this problem remains a
difficult generalization task in robotics --- in large part due to the lack of
methods for autonomously inventing \emph{
generalizable, abstract concepts and world models} that can be
transferred to more complex settings. Recent advances in robot learning address complementary problems
of learning tasks with short horizons (e.g., closing a
door)~\citep{hafner2023mastering} and of learning to imitate demonstrations in
scenarios with minimal differences from training
tasks (e.g., tying shoelaces or chopping fruits) ~\citep{fu2024mobile,brohan2022rt,zhaoaloha,black2024pi_0,barreiros2025careful} .

% Research in psychology and computational cognitive science indicates that humans learn and make extensive use of  
%  abstract models for planning~\citep{rosenbaum1983hierarchical,botvinick2008hierarchical,dezfouli2013actions,solway2014optimal,ho2022cognitive,ho2022people}.

% This paper presents the first approach for...

\begin{figure*}
% \begin{wrapfigure}[17]{l}{0.51\textwidth}
% \vspace{-1.5em}
\centering
    \includegraphics[width=0.9\textwidth]{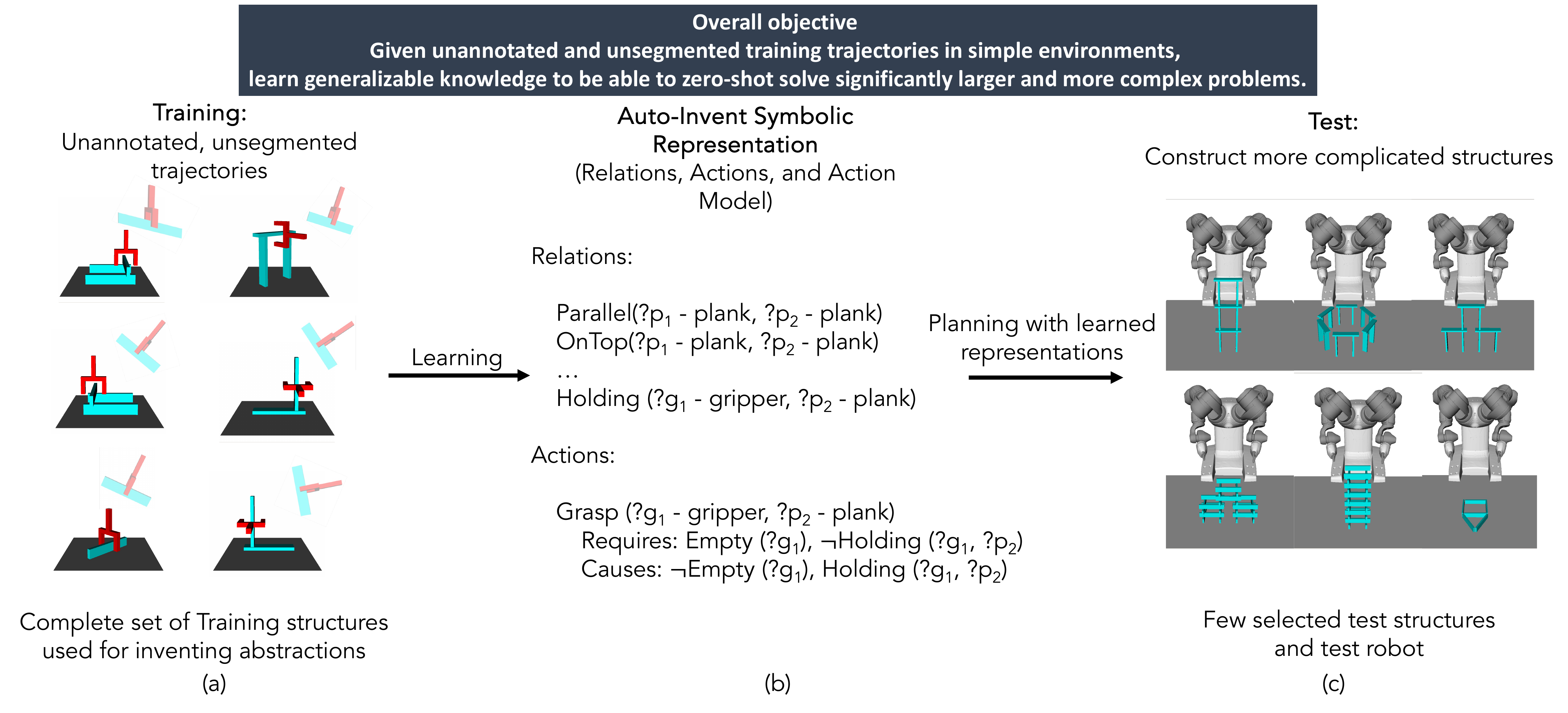}
    \caption{ Example of transfer achieved in this paper. (a) shows the complete set of training tasks for one of our experiment domains; (b) shows the automatically invented symbolic concepts and world models by our approach; (c) shows some of the test tasks zero-shot solved using the learned model despite significantly greater numbers and complexity. Interpretable relation names are provided for clarity, not learned by the method.}
    \label{fig:overall}
    % \vspace{-1em}
% \end{wrapfigure}
\end{figure*}

\paragraph{Core contribution} This paper presents the first approach for learning to invent abstract logic-based concepts and world models
from raw demonstrations in the form of of kinematic trajectories of the robot without any human annotation,
segmenting, or guidance about primitive controllers, actions, or concepts. Fig.~\ref{fig:overall} shows an overview of our approach. 
Our input is a small set of raw training trajectories over kinematic states from simple tasks. The output is a set of auto-invented logical concepts with concrete semantic definitions, a set of high-level actions with auto-generated pose-generators for refinement, and a logic-based world model. No human annotation is required in the entire process. Extensive empirical analysis in a range of mobile manipulation settings in simulation and in the real world show that the learned concepts enable the robot to solve tasks it has never encountered before, with up to 18$\times$ in the number of objects that need to be manipulated, novel goal configurations and significantly larger horizons. 

{\textbf{Related work} \hspace{1em}} Typically, logic-based concepts are hand-crafted by human experts~\citep{shah2020anytime,garrett2021integrated}. This requires extensive domain engineering efforts limiting the applicability of autonomous robots to tasks envisaged prior to deployment. 
  Other related work shows that unary predicates and actions can be learned for motion planning~\citep{shah2022using,shah2024hierarchical} but not for mobile manipulation. Foundation models have been used for composing high-level robot skills provided as Python APIs~\citep{liang2023code}, for translating natural language instructions to logic-based formulas~\citep{liu2023llm+}, and for planning~\citep{driess2023palme,brohan2023can}. These methods rely on expert-provided knowledge about high-level robot actions and the procedures for executing these actions. 
Recent work indicates that logical concepts are learnable given a-priori knowledge of high-level robot 
skills or actions~\citep{konidaris2018skills}, and that more complex concepts can be learned from an input set of concepts~\citep{silver2022learning,silver2023PredicateInvention}. However, the problem of learning logical concepts and actions without prior knowledge of either kind has remained understudied at best. We show for the first time that
it is possible to solve this problem, and to learn symbolic world models that support complex reasoning for solving significantly larger tasks than the ones that are encountered in training demonstrations.
%entirely from raw demonstrations  (Fig.~\ref{fig:overall}). 
Detailed comparison with other related work is presented in Appendix~\ref{app:related_work} .

%===============================================================================

\section{Problem Setting}
\label{sec:problem_setting}
We represent the set of possible  configurations as $\gX$, where each state $x\in\gX$ specifies 
6-dimensional poses (positions and orientations) of objects in the environment,
links of the robot, and the set of objects attached to the
robot.  The set of \emph{primitive actions} ($\gA$) represents actions that induce minor changes in the robot pose that
are small enough for a native robot controller to execute. Such actions are
typically executed in under a second and affect pose changes of at most a few
millimeters.
% Native robot actions
% change the pose of the robot and of the objects attached with robots.
% % However, these changes are small such that a robot controller can execute them in a single step. 
% Typically, these primitive actions are executed by the robot in less than $1$
% second and do not change the pose of the robot (and the objects) by more than a
% few millimeters. 
However, there are uncountably many infinite primitive actions, many 
of which transform the configuration space (such as closing
the gripper around an object to obtain a stable grasp); tasks of interest in
this paper comprise of composing tens of thousands of such actions. This makes na\"ive planning via optimization over all possible sequences of
primitive actions infeasible.  
Formally, this planning problem  can be expressed as an optimization of a utility function over the space of all possible policies as follows: 
\[ \underset{\pi} \argmax\; \{ J\left( \pi,x_0, H  \right) \text{ such that } \pi: \gX \rightarrow \gA    \text{; } J \text{ is a utility function} \} \] 
Here, we use a satisficing formulation for $J$ that assigns a positive utility to policies meeting a goal condition $\gG$ within $H$ time steps. However, the same method is valid for optimization models where $J$ accounts for action costs. Solving this optimization problem in a deterministic setting would yield a motion plan i.e., a sequence of primitive actions starting from $x_0$ and reaching a state that fulfills the goal condition $\gG$.

% argmax {J(\pi(s_o, H))} over \pi's \in \mathcal{F: {c\rightarrow a} c\in \X: config space, a\in A^P: primitive actions} J is the utility function. We use a satisficing formulation here  that associates a positive utility with policies that achieve a goal condition \mathcal{G} in H time steps, but the same framework applies to optimizing formulations where J reflects action costs  

% and the problems that we intend
% our robots to tackle require up to tens of thousands of such actions. Therefore,
% na\"ively using these primitive actions in infeasible. 

\paragraph{Abstract world models} Since solving the planning
problem at the level of detail of primitive actions is intractable even for simple tasks, the conventional approach is to use hand-crafted abstract logical concepts and  create abstract  world models for solving multiple tasks in a domain.
These concepts include state
abstractions in terms of relationships such as ``On(obj, table)'' and temporal
abstractions or conceptual, high-level actions such as ``pickup(obj)'' and
``place(obj, loc)''. Thus world models consist of three  components: a
\emph{logical concept vocabulary} $\mathcal{V}$ with semantics (functions determining when each predicate is true), a set of \emph{high-level
actions} $\bar{\gA}$, and \emph{action interpreters}, or methods for
translating high-level actions into a robot's primitive actions. Each high-level
action in $\bar{\gA}$ is defined using a precondition (a logical formula over
$\mathcal{V}$ describing when the action can be executed) and an effect (a
logical formula over $\mathcal{V}$ describing the effects of the action). 
% E.g.,
% the action \texttt{Pickup(Robot, Object)} can be executed if  the robot's
% gripper is empty and results in the object being in the gripper.
Each high-level
action needs action interpreters that ``refine'' it into motion plans that can
be executed by the robot. Such world models represent knowledge about the world
while abstracting away the specifics of robot and object poses, which may change
with changes in goals and problem situations~\citep{McDermott_1998_PDDL}. This yields significant
generalizability, as the conceptual relations and actions are task agnostic.

The use of abstract concepts changes the formal planning problem to an optimization over abstract policies defined as follows: 
\[ \underset{[\pi]}\argmax\; \{ J(\pi,[x_0],H) \text{ such that } [x_0] = \alpha(x_0) \text{ and } [\pi]: \alpha(\gX) \rightarrow \beta(\gA) \text{ and } \alpha, \beta \text{ satisfy } \kappa \}
\]
Here, $\alpha$ and $\beta$ are hand-coded state and action abstractions respectively that satisfy refinability constraints $\kappa$ often crafted by expert roboticists based on their
experience with the expected tasks and domains of deployment. This is the task and motion planning (TMP) setting~\citep{cambon2009hybrid,garrett2021integrated,srivastava2014combined,shah2020anytime} (described more in App.~\ref{app:environemnt}). However, this formulation introduces dependency on hand-crafted knowledge, and it can result in incorrect solutions because it is difficult to design $\alpha$ and $\beta$ that satisfy refinability. 

\paragraph{Critical regions}
 Our approach builds on the notion of \emph{critical regions}~\citep{molina2020learn,shah2022using}, which generalizes and unifies the concepts of hubs and bottlenecks.
Earlier work defines critical regions in a goal-agnostic manner, however, in this work we consider goal-conditioned critical regions. Intuitively, as the name suggests, goal-conditioned critical regions learn critical regions for a specific training problem. We learn goal-conditioned critical regions for each training task and combine them in order to compute the set of critical regions. Given a robot with a configuration space $\gX$, goal-conditioned regions are defined as follows. 

\begin{definition}
    \label{def:cr} 
    Given a set of solutions for a robot planning problem $T$, the measure of criticality of a Lebesgue-measurable open set $\rho \subseteq \gX$, $\mu(\rho)$, is defined as $\lim_{s_n \to ^{+}\rho} \frac{f(\rho)}{v(s_n)}$ where $f(r)$ is a fraction of observed motion plans solving the task $T$ that pass through $s_n$, $v(s_n)$ is the measure of $s_n$ under a reference density (usually uniform), and $\to^{+}$ denotes the limit from above along any sequence $\{s_n\}$ of sets containing $\rho$ ($\rho \subseteq s_n$, $\forall\,n$). 
\end{definition}
% We now describe our approach for automatically learning state and action abstractions. 

\begin{figure*}[t!]
  \centering
  \includegraphics[width=0.85\textwidth]{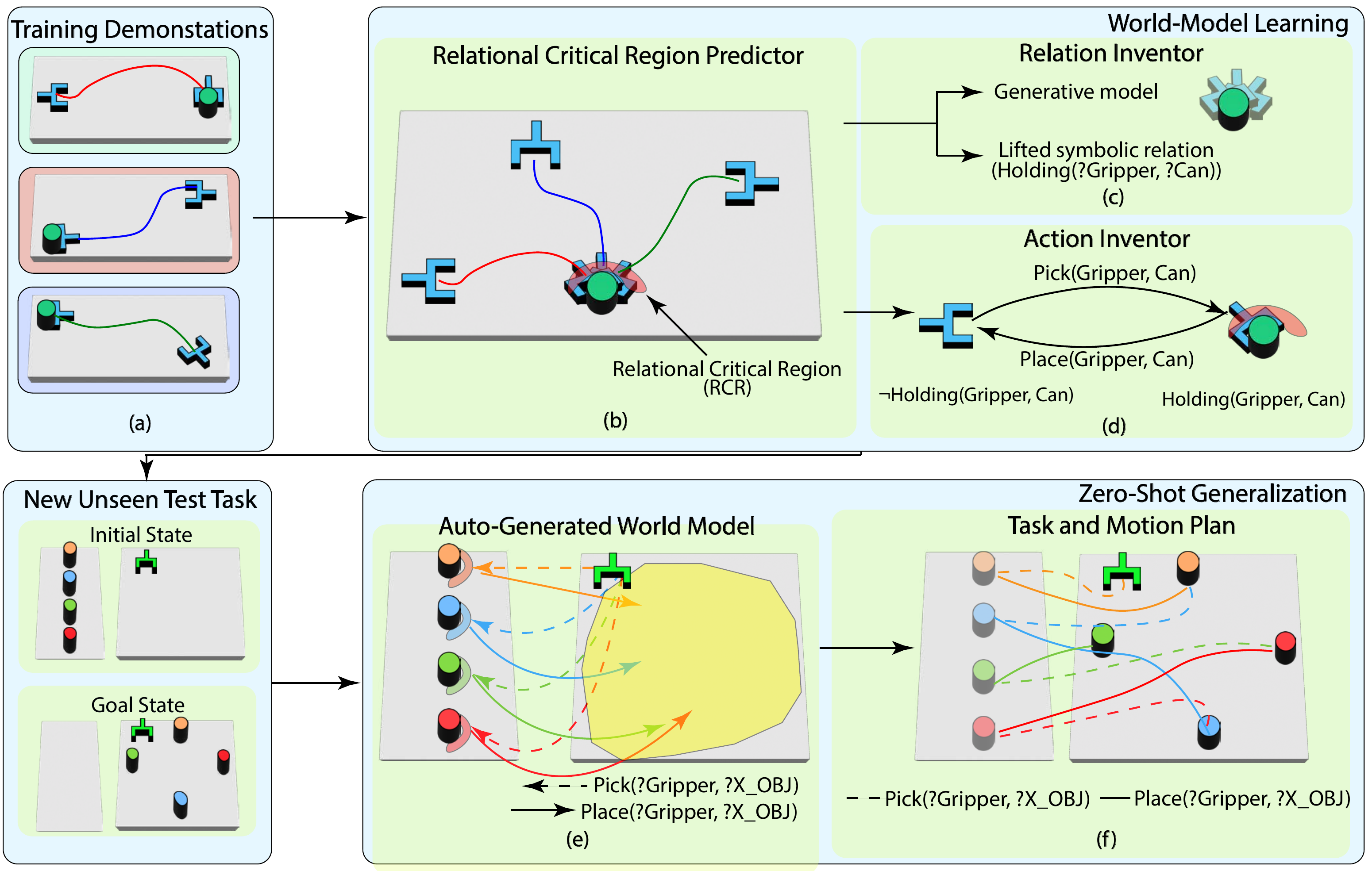}
  \caption{ Overview of LAMP. From unlabeled, unsegmented demonstrations (a), LAMP learns relational critical region predictors (b), each defining relations between object pairs (c). These relations form abstract states, with learned actions as transitions (d), together constituting a symbolic world model learned from scratch. For new tasks with more objects or obstructions, LAMP predicts new relations to build the abstract state space and goal, then uses the model to synthesize behaviors that achieve it (f).}
  \label{fig:img2}
  % \vspace{-1em}
\end{figure*}

\section{Our Approach} 
% This paper uses a set of raw training demonstraitons

% Our central idea is to learn abstract symbolic concepts and high-level world models that can be transferred to new test tasks with unseen goals.
% In this context, different goals correspond to different numbers and configurations of the objects in the goal state (e.g., strucutres with different numbers of planks and configurations in Fig.~\ref{fig:overall}). We formally define the abstraction learning problem as follows. 

% \begin{definition}
% \label{def:problem}

% Let $T_\emph{train}$ be a set of training tasks and $\gD_\emph{train}$ be a set of demonstrations that successfully solve the training problems. We define the \textbf{abstraction learning problem} as learning $1)$ a set of generative symbolic concepts $\gV$, $2)$ a set of generators $\Gamma$ for each invented symbolic concept, and  $3)$ a set of high-level actions $\bar{\gA}$.
% \end{definition}

The objective of this work is to enable robots to create well-founded conceptual world models by inventing abstract concepts. 
%We investigate whether the resulting system can enable robots to perform tasks beyond the demonstrations, reaching the level of generality typically achieved with hand-crafted abstractions.
It turns out that by including the abstraction as a parameter in the optimization objective, the agent can learn to perform better as it no longer depends on human-crafted abstractions. Formally, this  changes the problem to:
\[
\underset{\pi,\alpha, \beta} \argmax \; \{ J(\pi, x_0, H) \text{ such that } \pi : \alpha(\gX) \rightarrow \beta(\gA)  \text{ and } \alpha,\beta \text{ satisfy } 
\kappa \}
\]
where the abstractions become a part of the parameters to be optimized. 
%We found that these abstractions can be automatically learned  from a limited set of demonstrations without the need for error-prone hand-labeling, prompting, or annotation.

% The core contribution of this paper is the first known approach for simultaneously 
% inventing a predicate vocabulary and abstract actions that solve unseen test problems
% $T_\emph{test}$ with similar robots and types of objects but significantly varying goals.
 Our approach -{}- \emph{Learning Abstract Models for Planning (\textbf{LAMP})} -{}-  automatically learns these abstractions from kinematic demonstrations and represents them as world models in the planning domain definition language (PDDL~\cite{McDermott_1998_PDDL}).
 LAMP consists of the following steps ( App.~\ref{app:overall}): $(i)$ learn to predict a general, relational form of critical regions called \emph{relational critical regions} that characterize  salient regions in the relative state spaces among objects (Sec.~\ref{subsec:proto_relations}), $(ii)$  invent a symbolic relational vocabulary  based on predicted relational critical regions (Sec.~\ref{subsec:relational_vocab}), and $(iii)$ invent high-level actions and learn symbolic world models (Sec.~\ref{subusec:learning_actions}). The resulting knowledge, i.e. a symbolic world model, is then used to zero-shot solve new tasks. 
 % in zero-shot mode.  
 We now present  these components in detail.

% Problem statement

% overview of next few sections [2-3 sentences]

\subsection{Learning Proto-Relations}
\label{subsec:proto_relations}

We postulate that high-level robot actions currently crafted by experts are effectively transitions to and from salient regions within the environment and that such regions can be automatically discerned. 
%A region is deemed salient if it is essential for the completion of a specified task. 
E.g., Fig.~\ref{fig:img2} illustrates a task where the gripper needs to pick up a can. The area from which the can can be grasped constitutes a salient region.  Classically hand-crafted ``Pick" and ``Place" actions transition the gripper into and out of this salient region, respectively. Thus, if we automatically identify such salient regions, high-level actions can also be invented autonomously, subsequently enabling autonomous formulations of generalizable world models for long-horizon problems.

% \begin{figure*}[t!]
%   \includegraphics[width=\textwidth]{nature_images/relations.pdf}
%   \caption{Different relations invented by our approach and their corresponding critical regions. Each image shows one binary predicate and its semantic interpretation. The red dots show sampled possible poses for the object in the relational critical region. }
%   \label{fig:relations}
% \end{figure*}

\paragraph{Relational critical regions}

% \begin{wrapfigure}[18]{l}{0.5\textwidth}

% \end{wrapfigure}

We formalize the notion of salient regions as \emph{relational critical regions (RCRs)}. Critical regions (CRs, Def.~\ref{def:cr})  help identify these regions in the robot's configuration space~\citep{lavalle2006planning}. However, for long-horizon planning involving multiple objects, salient regions occur not in the robot's absolute configuration space but in the space of relative poses among objects. E.g., in Fig~\ref{fig:img2}(a) grasping poses for the can constitute a relational critical region (shaded red in Fig.~\ref{fig:img2}(b)), regardless of the can's location.  Intuitively, a relational critical region for object $x$ w.r.t. object $y$ is a region in the set of poses of $x$ relative to $y$ that has a high density of solutions for a given distribution of tasks. We thus generalize critical regions to define \emph{relational critical regions} as follows.
\begin{definition}
    \label{def:rcr}
    Let $T$ be a robot planning problem and $\gD_T$ be a set of solution trajectories for the planning problem T. 
    Let $o_1, o_2 \in \gO$ be a pair of objects, and let $\gX^{o_1}_{o_2}$ define the set of relative poses for object $o_2$ in the frame of $o_1$. The measure of the criticality of a Lebesgue-measurable open set $\rho \subseteq \gX^{o_1}_{o_2}$, $\mu(\rho)$, is defined as $\lim_{s_n \to ^{+}\rho} \frac{f(\rho)}{v(s_n)}$ where $f(\rho)$ is the fraction of observed solution trajectories solving for the planning problem $T$ that contains a relative pose $P^{o_1}_{o_2}$ such that, $P^{o_1}_{o_2} \in \rho$ $v(s_n)$ is the measure of $s_n$ under a reference density (usually uniform); and $\to^{+}$ denotes the limit from above along any sequence $\{s_n\}$ of sets containing $\rho$ ($\rho \subseteq s_n$, $\forall\,n$). A region $\rho \subseteq \gX_{o_2}^{o_1}$ is a \emph{relational critical region} (RCR) iff $\mu(\rho)$ is greater than a predefined threshold $\theta$. 
\end{definition}

LAMP takes demonstrations of simple tasks in the form of kinematic-state trajectories and converts them into trajectories of relative poses between object pairs. These trajectories are clustered into pairwise occupancy matrices between object-type pairs $(\tau_i,\tau_j)$, and Def.~\ref{def:rcr} is applied to identify clusters with high criticality. LAMP then applies the \texttt{label}\footnote{\url{https://github.com/opencv/opencv-python}} function from the OpenCV package to extract connected components from each occupancy matrix. Let $\Gamma$ denote the set of all connected components, and $\Gamma_{ij} \subset \Gamma$ the subset associated with object-type pair $(\tau_i,\tau_j)$.

From each connected component $\gamma \in \Gamma$, LAMP uniformly samples poses and fits a Gaussian mixture model (GMM) using \texttt{GaussianMixture}\footnote{\url{https://scikit-learn.org/1.5/modules/mixture.html}} from scikit-learn, parameterized by $(\mu_\gamma, \Sigma_\gamma)$. The number of Gaussian components is chosen according to the number of connected components within $\gamma$. Each resulting mixture model defines a relational critical region (RCR) $\psi_\gamma$. A relative pose $p_{o_1}^{o_2}$ is classified into $\psi_\gamma$ (i.e., $\psi_\gamma(p_{o_1}^{o_2}) = 1$) iff its likelihood under $(\mu_\gamma,\Sigma_\gamma)$ exceeds a threshold $\epsilon$. For brevity, we omit the subscript $\gamma$ when clear from context.

Once trained, these mixture models act as generative predictors, enabling zero-shot inference of relational critical regions in novel environments, given the object configurations and occupancy matrix of the environment.

% Using Def.~\ref{def:rcr}, these trajectories are clustered into relational critical regions (RCRs) $\Psi$ for each object pair $o_i, o_j$.

% For each object-type pair $(\tau_i,\tau_j)$, let $\Psi_{ij} \subset \Psi$ denote the corresponding RCR clusters. LAMP fits a Gaussian mixture model (GMM) to the poses within $\Psi_{ij}$: each RCR $\psi \in \Psi_{ij}$ is represented as a Gaussian component with parameters $(\mu_\psi, \Sigma_\psi)$. A relative pose $p_{o_1}^{o_2}$ is assigned to cluster $\psi$ (written $p_{o_1}^{o_2} \in \psi$) iff its likelihood under $(\mu_\psi, \Sigma_\psi)$ exceeds a threshold $\epsilon$ (denoted as $\psi(p_{o_1}^{o_2}) = 1$).

% In practice, we first compute an occupancy matrix using training demonstrations, and then use the \texttt{label} function from the \texttt{OpenCV} library\footnote{\url{https://github.com/opencv/opencv-python}}
 % to extract connected components from it. Each connected component is treated as an RCR, and number of components in each connected component is used as the number of Gaussian components for the GMM\footnote{\url{https://scikit-learn.org/1.5/modules/mixture.html}}. Training data for the mixture model is generated automatically by sampling poses uniformly from each identified RCR cluster. Once trained, these mixture models serve as generative predictors that can zero-shot predict relational critical regions in unseen environments, given the object configurations and occupancy matrix of the new environment.

We now discuss our approach for learning symbolic relational vocabulary from the predicted RCRs. 

\subsection{Inventing Semantically Well-Founded Concepts for Logic}
\label{subsec:relational_vocab}

Relational critical regions represent salient regions in the environment. However, they are insufficient for generalizable long-horizon reasoning; high-level reasoning requires abstract actions, and a relational vocabulary for expressing the pre- and post-conditions of these actions. Therefore, for each relational critical region predicted by the learned multivariate Gaussian predictors between a pair of objects,  LAMP's \emph{Relation Inventor}  (Alg.~\ref{alg:relation_inventor}, App.~\ref{app:relation_inventor})  creates a unique binary relation between the corresponding object types. E.g., the RCR shown in Fig.~\ref{fig:img2}(a) constitutes the extent of a newly invented relation  (a concept equivalent to \texttt{Holding(Gripper, Can)}), and it is true when the gripper is in the shaded red region around the can.

% \begin{figure*}[t!]
% \centering
%   \includegraphics[width=0.9\textwidth]{nature_images/train_test_2.pdf}
%   \caption{Generalization across different tasks. (a) shows the training tasks used to learn relational region predictors and (b) shows the test tasks used to evaluate the overall method. We show that the predictors learned from very simple tasks can generalize to significantly complex test tasks that include more complex goals, more complex environments, and a larger number of objects (up to 18x). Note: training demonstrations for Keva and Jenga structures include same structures but different robots to match the gripper configuration of the robot used while testing.}
%   \label{fig:train_test}
% \end{figure*}

Formally, let $\gO_{\tau_i}$ and $\gO_{\tau_j}$ be the set of objects of type $\tau_i$ and $\tau_j$, respectively, and let $\Psi_{ij}$ be a set of RCR predictors between $\gO_{\tau_i}$ and $\gO_{\tau_j}$.
The Relation Inventor defines a unique binary relation $R^k_{ij}: \gO_{\tau_i} \times \gO_{\tau_j} \to \{\top, \bot \}$ for each relation region predictor $\psi^k \in \Psi_{ij}$ such that $R^k_{ij}(o_i, o_j) = \top$ iff for the relative pose $p_{o_i}^{o_j}$, $\psi^k(p_{o_i}^{o_j}) = 1$, where $p_{o_i}^{o_j}$ is the pose of the object $o_i$ relative to the $o_j$.

Next, the Relation Inventor (RI) defines two additional sets of Boolean relations. First, given the sets of objects $\gO_{\tau_i}$ and $\gO_{\tau_j}$, it defines a relation 
$R'_{ij}: \gO_{\tau_i} \times \gO_{\tau_j} \to \{\top, \bot\}$ 
such that $R'_{ij}(o_i, o_j) \iff \forall k \left[ \neg R^k_{ij}(o_i, o_j) \right] $. 
Second, it defines a relation for each relational critical region predictor indicating whether multiple objects can occupy that RCR or not.  Intuitively, this models the free volume in the predicted region.
E.g., the RCR for gripper  w.r.t. can (Fig.~\ref{fig:img2}(b), red)  can be only occupied by a single can but the RCR for the placed cans w.r.t. table (Fig.~\ref{fig:img2}(e), yellow) can be occupied by multiple cans. 
Formally, given the sets of objects $\gO_{\tau_i}$ and $\gO_{\tau_j}$, 
% a set of 
relational regions predictors $\Psi_{ij}$, the RI defines a Boolean relation for each relational region predictor $\psi^k \in \Psi_{ij}$, $R_{ij}^{\emph{free}_k}: \gO_{\tau_i} \to \{\top, \bot\}$ such that it is true iff for  $o_i \in \gO_{\tau_i}$ and $o_j \in \gO_{\tau_j}$, $\rho_\emph{free}(\psi^k,o_i) > \rho(o_j)$. 
Here, $\rho_\emph{free}(\psi^k)$  is the free volume of the predicted region $\psi^k$ and $\rho(o_j)$, the volume of the object $o_j$.

% are automatically 
% learned by the Relation Inventor in the form of  generative  for the learned logical concept vocabulary.
Given a new task $T$ and its set of objects $\gO_T$, LAMP uses the automatically learned relational region predictors to predict the RCRs for objects and  generate the relational vocabulary $\gV_{T}$ for the new task.
% This vocabulary now can be used to represent each configuration $x \in \gX_T$ as a high-level state as a set of true relations given the configuration $x$.

We now discuss our approach for autonomously inventing symbolic actions and world models.

\subsection{Learning High-Level Actions and Logical World Models from Raw Data}
\label{subusec:learning_actions}
% \begin{wrapfigure}[21]{l}{0.5\textwidth}
%     \vspace{-1.5em}

% \end{wrapfigure}
The last step in the overall LAMP algorithm is to synthesize generalizable and transferrable actions, models, and action interpreters.
% We use the automatically learned relational vocabulary to invent these high-level actions.
Recall that we hypothesized that high-level actions are transitions to and from RCRs. E.g., the transitions in and out from the red RCR in Fig.~\ref{fig:img2}(a) induce high-level actions \texttt{Pickup(Gripper, Object)} and \texttt{Place(Gripper, Object)} respectively. 
Therefore, LAMP's \emph{High-Level Action Inventor (HLAI)} (App.~\ref{app:action_inventor}) use the predicted relational critical regions in training tasks to invent  high-level actions.
% using the \emph{Action Inventor} (Alg.~\ref{alg:action_inventor}, App.~\ref{app:action_inventor}). 
% Given the automatically learned relational region predictors and the predicted relational critical regions using these predictors, we can invent such high-level actions for an unseen test task. 
However, using the exhaustive enumeration of the set of predicted regions may lead to a large number of actions, most of which may be infeasible to realize. 
Instead, HLAI creates high-level actions corresponding to transitions between abstract states in the input training demonstrations, ensuring  that all actions are executable by the robot. 

More precisely, HLAI uses the invented relational vocabulary $\gV$ to first convert each state trajectory $\langle x_0, \dots x_n \rangle$ of a demonstration to abstract states $\langle s'_0, \dots, s'_n \rangle$, and then lifts it to lifted states $\langle s_0, \dots, s_n \rangle$ by replacing object identities with placeholder variables of the original object types. Given the relational vocabulary $\gV$ invented by RI (Sec.~\ref{subsec:relational_vocab}), each abstract state $s'_i = \{ R^k(o_i, o_j) | \forall o_i, o_j \in \gO, \forall R^k \in \gV, x \models R^k(o_i, o_j) \}$.

Next, for each lifted transition $C_{ij}  = s_i \to s_j$, HLAI computes sets of added and deleted relations such that $C_{ij}^+ = s_j \setminus s_i$ and $C_{ij}^{-} = s_i \setminus s_j$ and identifies the set of all transitions from the training demonstrations that induce the same $\langle C_{ij}^{+}, C_{ij}^{-} \rangle$.  Each of these sets induces a high-level action $\bar{a}_{ij}$. It repeats this process for every identified set and build the set of high-level actions $\bar{\gA}$. 
HLAI then uses associative action model learning to learn symbolic models for these actions (App.~\ref{app:action_inventor}).

\begin{wrapfigure}[18]{l}{0.52\textwidth}
% \begin{figure*}
\centering
\includegraphics[width=0.5\textwidth]{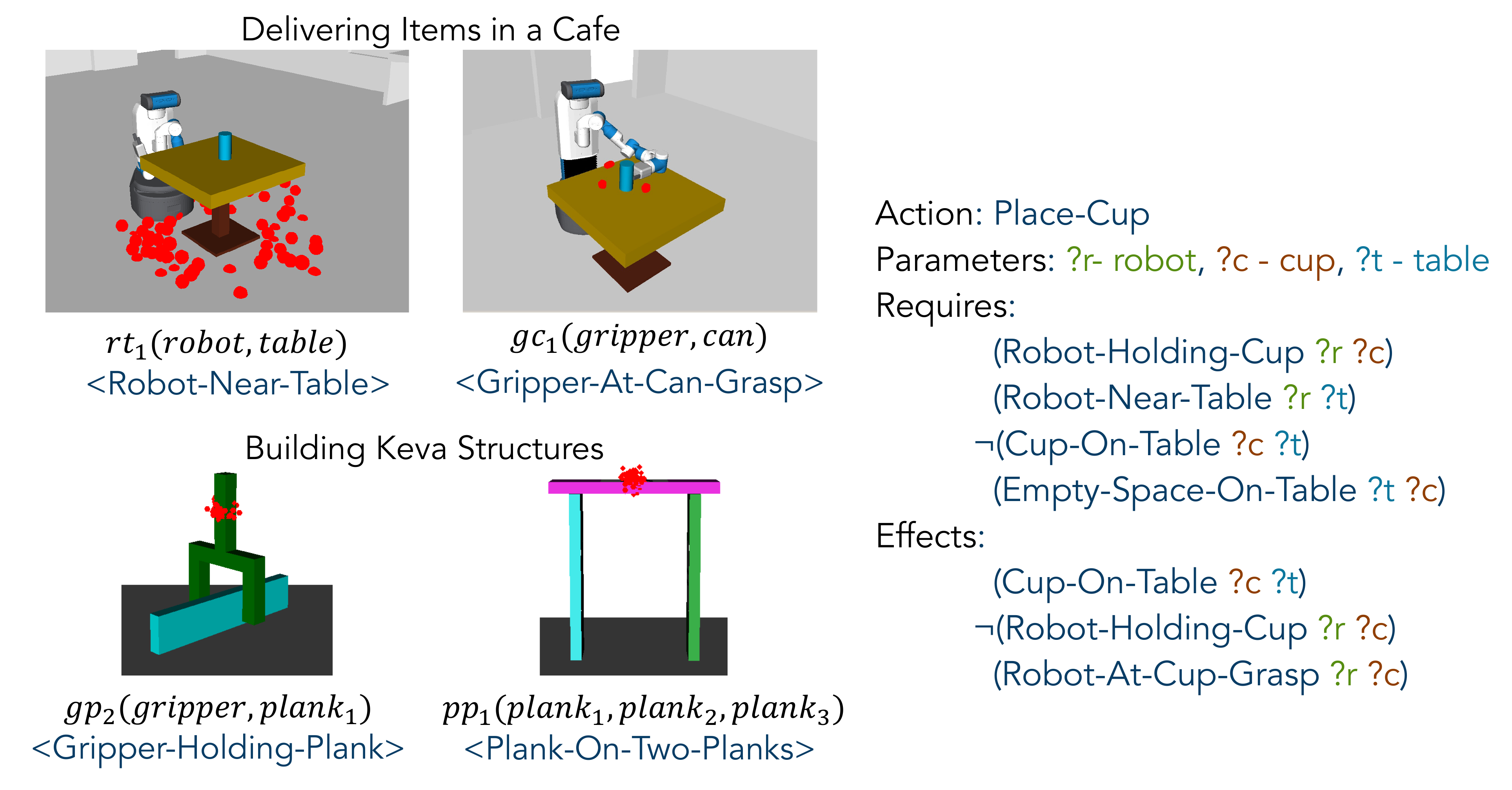}
  \caption{ Examples of relations and an invented action with their critical regions. (a) Each image shows a binary predicate with its semantic interpretation; red dots indicate sampled poses within the relational critical region. (b) An auto-invented action is shown with parameters, preconditions, and effects. Relation names in blue reflect author-provided interpretations.}
  \label{fig:relations}
  % \vspace{-1.5em}
% \end{figure*}
\end{wrapfigure}
Each action invented in this manner can have spurious preconditions corresponding to static relations that do not change when the action is applied, but hold true in states where the action is executed. Therefore, the AI removes relations from the learned precondition that
(i) are not parameterized by any of the objects that are changed by the action and
(ii) that are not changed at any point in any of the demonstrations. 
This removes spurious preconditions with respect to the observed data. 

To execute learned symbolic actions, robots require interpreters that convert abstract actions into sequences of primitive actions. LAMP autonomously constructs these interpreters using learned multivariate Gaussian predictors, whose generative properties enable sampling feasible configurations. Formally, for a grounded symbolic relation $R'$, let \( \Gamma_{R'} = \{ x \in \gX \mid R'(x)=1 \} \) denote its interpretation region. Then, the interpreter for an abstract action $\bar{a}'$ with effects \( R'^1 \land \dots \land R'^k \) is defined as \( \Gamma_{\bar{a'}} = \Gamma_{R'^1} \cap \dots \cap \Gamma_{R'^k} \). 
% To execute $\bar{a}'$, a planner~\citep{shah2020anytime} samples from \( \Gamma_{\bar{a'}} \) and employs a motion planner to generate a suitable sequence of primitive robot actions. 
We build on  STAMP~\citep{shah2020anytime} for task and motion planning using learned representations. We mitigate inaccuracies in learning using predicate abstraction~\citep{graf1997construction} and a top-$k$ planner~\citep{katz-et-al-icaps2018}. App.~\ref{app:tamp} discusses this in detail.

% Lastly, LAMP automatically learns action interpreters in the form of generative multivariate Gaussian predictors that allow sampling relative states for corresponding objects. 

% Once the precondition and effect of an action are learned, the final step is to learn the parameters of the action that can be replaced with objects in order to ground the action. In this step, the relations in precondition and effect are processed in order. These relations are processed in alphanumeric order and each of their parameters is added to the action's parameter list, if not already added. This process leads to an ordered list of parameters of the action, which can be grounded with compatible objects.

\section{Empirical Evaluation}

% We designed five realistic settings to evaluate this approach in both simulated
% and real-world environments. These settings are: $(i)$ packing cans
% into a small box (\textbf{Box Packing}, with 200 demonstrations, with one can); $(ii)$ the assembly of complex free-standing structures using Keva planks (\textbf{Keva}, with 160 demonstrations with a maximum of three planks), 
% and $(iii)$ Jenga planks (\textbf{Jenga}, with 160 demonstrations with a maximum of three planks) with the YuMi and Fetch robots respectively; $(iv)$ autonomous food delivery in a
% café environment (\textbf{Café}, with 200 demonstrations with one item), and  $(v)$ setting 
% up a dining table with cups and bowls (\textbf{Dinner Table}, with 200 demonstrations with one cup or bowl). For all experiments
% in this work, LAMP used a maximum of 200 demonstrations on significantly simpler tasks, \emph{with only half of the
% demonstrations in each domain completing those tasks}. 

We evaluated LAMP across five realistic settings in simulation and the real world: $(i)$ packing cans into a small box (\textbf{Box Packing}, 200 demos, one can); $(ii)$ assembling free-standing structures with Keva planks (\textbf{Keva}, 160 demos, up to three planks); $(iii)$ stacking Jenga planks (\textbf{Jenga}, 160 demos, up to three planks) using YuMi and Fetch robots; $(iv)$ autonomous food delivery in a café (\textbf{Café}, 200 demos, one item); and $(v)$ setting a dining table with cups and bowls (\textbf{Dinner Table}, 200 demos, one object). Across all domains, LAMP trained on at most 200 demonstrations of simpler tasks--of which only half were successful.
% In all experiments, demonstrations are provided only in simulation and only for small
% problems (significantly smaller than test tasks). 

We show that the robot can invent concepts that generalize to larger, more complex tasks unseen during training—in simulation (Café, Keva, Box Packing) and in the real world (Dinner Table, Jenga). Figure~\ref{fig:relations} illustrates examples of these invented relations and a high-level action.
% The objective of this evaluation is to determine whether the robot can invent concepts 
\begin{figure*}[t]
	\centering
	\includegraphics[width=0.8\textwidth]{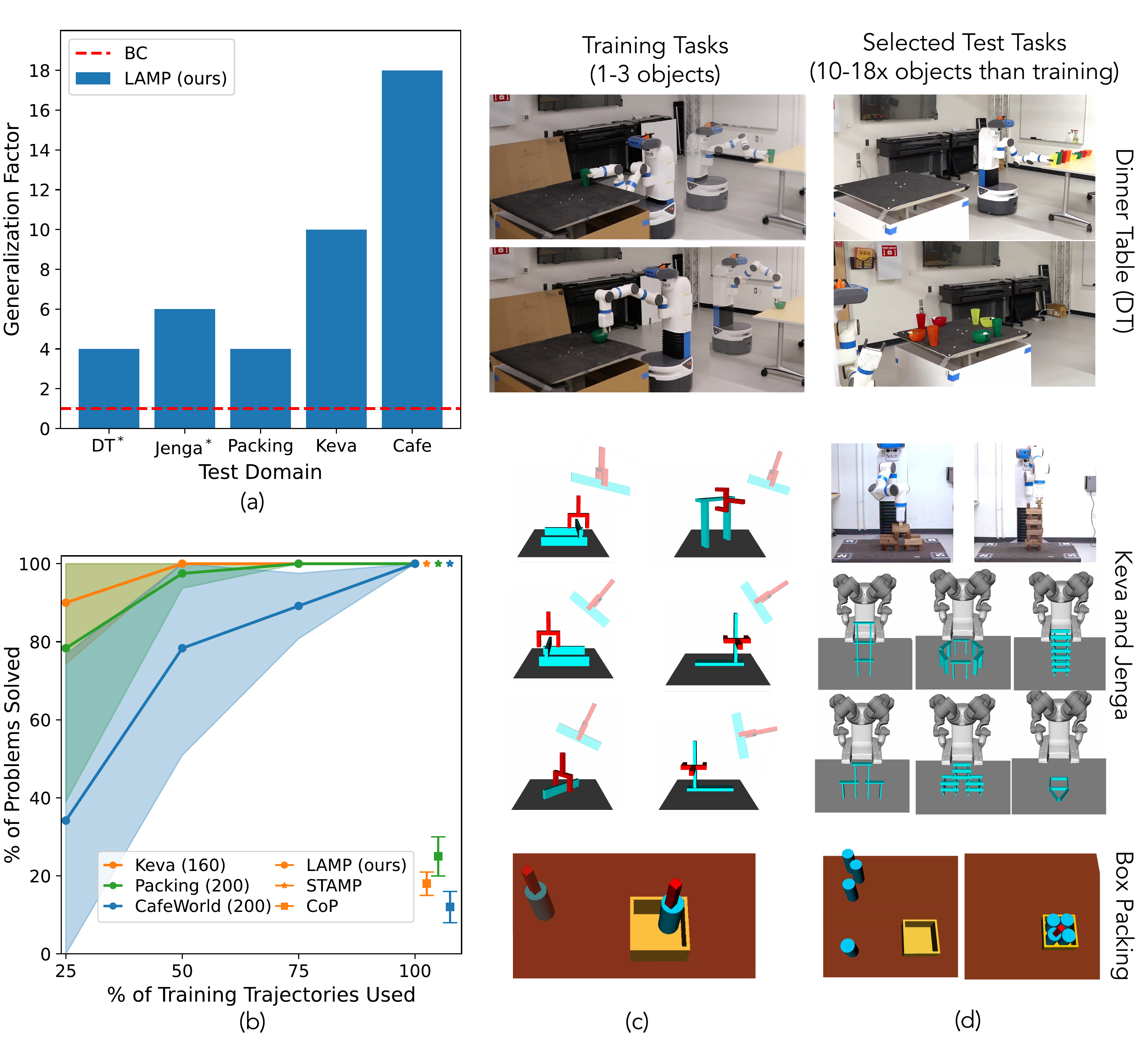}
	\caption{ Empirical evaluation of LAMP. (a) Generalization achieved by our approach, with the x-axis denoting domain and the y-axis the generalization factor. The red dotted line marks the 1× generalization zone typical of imitation learning or behavior cloning. (b) Robustness comparison with Code-as-Policies and STAMP, where the x-axis shows the number of training demonstrations and the y-axis the proportion of solved test tasks; shaded regions indicate standard deviation over 10 runs. (c) Training tasks used to learn symbolic world models. (d) Test tasks used to evaluate generalization.
	 }
	\label{fig:empirical}
	% \vspace{-1em}
  \end{figure*}
% to solve larger tasks either in the simulator (for the
% Café, Keva, and Box Packing tasks) or in the real world (for the Dinne Table
% and world models from  rudimentary tasks in simulation and use them
% table and Jenga structures). 
% Fig.~\ref{fig:relations}
% illustrates some of the concepts (relations and a high-level action) that the robot invented while solving these tasks.  

\paragraph{Goal specification} 
Our implementation  supports a variety of goal specification methods: $(i)$ In tasks where  goal poses are not anchored to a world-frame (e.g., Keva, Jenga) the system is given a goal structure that it automatically transcribes in terms of learned high-level predicates; $(ii)$  For tasks where goal poses are broader regions (e.g., Café, Box Packing) the goal is specified manually using invented predicates; $(iii)$ For tasks where 
the goal requires specific world-frame poses (e.g., Dinner Table) the goal is defined using  \texttt{AtGoal(x)} predicates which are defined to be true iff  object $\texttt{x}$ is in its goal location.

\paragraph{Learning beyond imitation: generalization factors} 
We define the \emph{generalization factor} as the ratio of the maximum number of
objects in test tasks to the maximum number of objects in training tasks. Since
the number of relevant objects characterizes the size of the state
space $\gX$ that the robot must plan over, this is an effective proxy for the complexity of a task; it is directly correlated
with the planning horizon, i.e., the number of decision-making steps that
the robot needs to consider. We used a maximum of three objects
in demonstrations across all test problems. 
% In all of our settings, we learn our relational world models
% in training environments with only 1-3 objects and evaluate their performance
% on tasks that involve significantly more objects and different settings than the
% training tasks.

 Fig.~\ref{fig:empirical}(a) shows the generalization factors for LAMP (blue bar) compared to the zone of generalization for contemporary approaches that use expert demonstrations for imitation (red line),
computed using 12 distinct test tasks for each domain with varying numbers of
objects or problem settings. Contemporary imitation learning approaches use expert demonstrations to learn to imitate and therefore only perform tasks that were trained on, leading to a generalization factor of 1. On the other hand, LAMP achieves a generalization factor of upto 18. In the café domain, LAMP was able to solve
problems with  18 (8) objects in simulation (real-worlxd) based on demonstrations
with 1 (simulated) or 2 (real-world) objects, yielding a generalization factor of
18$\times$ (4$\times$) in simulated (real-world) tasks.  Similarly, for
Keva and Jenga domains, LAMP invented its world models from
demonstrations with three planks and successfully built structures with 30 Keva
planks in simulated settings (generalization factor of $10\times$) using the
YuMi robot and 18 Jenga planks in real-world settings (generalization factor of
$6\times$) using the Fetch robot. The maximum possible generalization factor for
box packing was $4\times$ as the box can only accommodate four cans, and LAMP
achieved this generalization factor.

\paragraph{Zero-shot transfer: LAMP vs. symbolic and foundation model baselines}
% To evaluate the transferability of learning, we evaluate performance on new tasks starting  with no knowledge other than demonstration trajectories on simpler tasks with fewer objects (Fig.~\ref{fig:empirical}(b)). We compare LAMP's performance with a
% robot equipped with hand-crafted models, and a robot equipped with a foundation model-based reasoner
To evaluate transferability, we test LAMP on new tasks with no prior knowledge--using only demonstrations from simpler tasks with fewer objects (Fig.~\ref{fig:empirical}(b)--and compare its performance to robots with hand-crafted models and with foundation model–based reasoning~\citep{rana2023sayplan,yu2023language,liang2023code,driess2023palme}.
We used prior work on stochastic task and motion
planning (STAMP)~\cite{shah2020anytime} as a baseline approach for the former
and Code as Policies (CoP)~\cite{liang2023code} for the latter approach.
LAMP solved 100\% of the unseen test tasks in our evaluation reaching similar performance of STAMP, which used expert-provided abstractions. 
% On the other hand, CoP was able to solve only the simpler test tasks reaching less than 35\% success rate on overall evaluation. 
CoP solved only the simpler tasks, with under 35\% overall success.
We remark that although popular, these  baselines  solve fundamentally simpler problems as they require manually crafted inputs for the concepts our agent needs to invent on its own: STAMP requires complete world models, while CoP requires Python APIs for actions, interpreters, detailed goal descriptions with hand-crafted predicates, and sample solution code.
Additionally, CoP was given 25 attempts per problem with a
single successful execution required for the problem considered solved
(pass $@25$).

\paragraph{Sample efficiency and robustness} Sample efficiency is a critical
factor in robot learning. Modern approaches typically require tens of thousands
of demonstrations, which can be difficult to obtain due to the substantially higher cost of obtaining
demonstrations in robotics as compared to learning in other domains such as text and images.
% ~\citep{brohan2022rt,zhaoaloha}. 
We observed that enabling the robot to learn generalizable concepts
substantially reduces the burden of facilitating demonstrations. 
Figure~\ref{fig:empirical}(b) shows how this approach scales as the number of
demonstrations is reduced further, illustrating that LAMP can effectively
learn effective world models using as few as 40 goal-achieving  demonstrations.

\section{Conclusion}
This paper presents the first approach for  learning symbolic concepts and world models
directly from kinematic 
demonstration trajectories, enabling generalization across robots and unseen problem settings. Extensive evaluation in
simulated and real-world settings shows that the learned abstractions are
both efficient and interpretable.
Future work will leverage these autonomously learned abstractions to make robot re-tasking accessible to 
non-experts. We also aim to extend our approach to handle stochastic environments with stronger theoretical guarantees.

\clearpage
\section{Limitations}
% \nsalert{TODO: add limitations. Required by CoRL. Does not count towards pagelimit}

Despite the substantial gains in generalization and sample efficiency, further work is needed on problems not addressed in this work. First, the current formulation assumes near‐perfect  kinematic state estimation during training and deployment. Learning good state estimators is an active and orthogonal direction of research (e.g., using point cloud data~\citep{wen2024foundationpose}).
%However, the current available methods are noisy and unreliable rendering them inadequate for learning symbolic abstractions. 
We handle these limitations as follows: $(i)$ By using training data collected in a simulator with perfect state estimation, and $(ii)$ by using a motion capture system at test time. However, reliably bridging this gap will require tighter integration with robust state-estimation pipelines or explicit treatment of perceptual uncertainty inside the abstraction-learning loop.

Second, we restrict the scope of this paper to deterministic world models. However, most real-world settings are stochastic and require stochastic world models for safe and reliable planning~\citep{shah2020anytime}. This foundational work presents a step in the direction of learning more general world models for stochastic, and eventually, partially observable settings. 
% In the future, we aim to extend our approach to account for stochasticity in the environment while providing strong theoretical guarantees on learned world models. 

Third, in this paper, we limit our experiments of generalization to novel goals that use the same known object types. This includes scenarios with additional instances of these objects and with new positions and orientations. Future work will extend our approach to generalize learned concepts to entirely new object categories and novel geometries.

Finally, our approach is currently restricted to learning actions where dynamics do not play a significant role and the quasi-static assumption of motion planning is valid. Tasks requiring force-controlled contact, or closed-loop reactive policies lie outside the scope of the current work and constitute promising directions for research.

% In the future, we aim to extend our approach to learn such complex dynamics-based robot skills. 

% The paper focuses on learning 

% -- sim2real gap but require accurate poses but actively pursued can be noisy complimentary. 

% -- skills as E.g., DMPs tactile and dynamic feedback 

% -- non physical relations.. 

% \clearpage
% The acknowledgments are automatically included only in the final and preprint versions of the paper.
\acknowledgments{We thank the CoRL reviewers for their useful and constructive feedback. We thank Pulkit Verma for his help with the preliminary version of the presented work. We also thank Prof. George Konidaris with his helpful suggestions and discussions. The work was partially funded by NSF under the grants IIS 2451108 and IIS 1942856, and by ONR under the grant N00014-23-1-2416.}

%===============================================================================

% no \bibliographystyle is required, since the corl style is automatically used.
\bibliography{ref}  % .bib

\clearpage

\section*{Appendix}
\renewcommand{\thesubsection}{\thesection}
\renewcommand\thesection{\Alph{subsection}}

\subsection{Formal Framework}
\label{app:environemnt}

We consider a setting where the environment comprises objects and robots. 
Each object's state is represented by a 6D pose. A robot, considered a distinct object, is structured as a kinematic chain of links and joints, and its state is $\langle P_\text{base}, \Theta \rangle$, where $P_\text{base}$ indicates the 6D pose of the initial link, and $\Theta$ corresponds to the values for each joint. In an environment with objects $\gO = \langle o_1, \dots o_n, r_1, \dots r_m \rangle$, the state space is denoted by $\gX = \gX_{r_i} \times \gX_{o_j}$ for every robot $r_i \in \gO$ and object $o_j \in \gO$. A collision function $c$ categorizes the state space $\gX$ into two subsets: $\gX_\emph{free}$ (states without collisions) and $\gX_\emph{obs}$ (colliding states).

Primitive robot actions allow robots to alter their state, including configuration and base link pose, enabling movement and object manipulation within the environment. A primitive action $a$ is characterized by a deterministic function $a: x \mapsto x'$. Given environment states $x \in \gX$ and $x' \in \gX$, taking action $a$ in state $x$ leads to state $x'$. We describe a robot planning problem as follows:
\begin{definition}
 A \textbf{robot planning domain} is characterized as a tuple $\langle \gO, \gT, \gX, \gA, x_i, \gX_g \rangle$ where $\gO$ represents a collection of objects, $\gT$ is a set of object types, $\gX$ denotes a state space, and $\gA$ is an uncountably infinite set of deterministic native actions. The initial state of the environment, $x_i \in \gX_\emph{free}$ is an initial state of the environment, and $\gX_g \subseteq \gX$ is a set of goal states. Solving a planning problem involves finding a sequence of native actions $a_0,\dots,a_n$ such that $a_n(\dots(a_0(x_i))) \in \gX_g$. 
\end{definition}

Our approach extensively uses \emph{relative poses}. Every object in the environment also defines a frame of reference. A relative pose defines the pose of an object in the reference frame of another object. Basis transformations from linear algebra can be used to compute relative transformations of objects w.r.t to other objects in the environment. We refer to the pose of an object $o_1$ relative to an object $o_2$ as $P^{o_2}_{o_1}$. Let $\tilde{\gX}^{o_2}_{o_1}$  define a relative state-space for the pair of objects $o_1$ and $o_2$, i.e., the set of all poses of the object $o_1$ in the relative frame of the object $o_2$, 
and $\tilde{\gX}$ define the set of relative state spaces such that $\tilde{\gX} = \{ \tilde{\gX}^{o_i}_{o_j} | o_i, o_j \in \gO \land o_i \neq o_j \}$. 
Lastly, we define a transformation function $\xi: \gX \rightarrow \tilde{\gX}$ that computes the relative state for each absolute state of the environment.

\paragraph{Symbolic World Models}
We treat symbolic world models as first-order logic frameworks and use PDDL~~\citep{McDermott_1998_PDDL} to express these models. The PDDL domain encompasses two key components: $\langle \gV, \bar{\gA} \rangle$. Here, $\gV$ represents a collection of symbolic relationships, while $\bar{\gA}$ comprises high-level actions performed by robots.
The relations $R \in \gV$, defined by typed parameters, establish how objects of one type relate to one another type. These relations $R \in \gV$ can be instantiated with specific objects, termed $R$ when uninstantiated and $R'$ when instantiated. 
Instantiated relations $R'$ serve as Boolean classifiers, meaning they are true in a low-level state $x$ (indicated by $R'_x = 1$) if the relation is valid for the objects in question within the state $x$. 
The abstraction function $\alpha: x \mapsto s$ evaluates all instantiated relations in a low-level state $x \in \gX$, resulting in an abstract grounded state $s' \in 2^{\gV'}$ (or $s \in \alpha(\gX)$). This symbolic grounded state $s'$ comprises the relations that hold true in low-level state $x$, whereas the symbolic lifted state is denoted by $s$ ($s \in 2^{\gV}$). 

 $\bar{\gA}$ outlines symbolic lifted actions using lifted relations $\gV$. Each action $\bar{a} \in \bar{\gA}$ has typed symbolic parameters. $\bar{a}$ is defined as a tuple $\langle \emph{pre}_{\bar{a}}, \emph{eff}_{\bar{a}} \rangle$, where $\emph{pre}_{\bar{a}}$ is a conjunctive formula of parameterized relations $\gV$. The action's effect $\emph{eff}_a$ is a tuple $\emph{eff}_{\bar{a}} = \langle \emph{add}_{\bar{a}}, \emph{del}_{\bar{a}} \rangle$ adding relations $\emph{add}_{\bar{a}}$ and removing relations $\emph{del}_{\bar{a}}$ from the state once the action $\bar{a}$ is executed. Actions $\bar{a}$ can be grounded to specific objects, resulting in grounded actions $\bar{a}'$, generating grounded precondition $\emph{pre}_{\bar{a}'}$ and effect $\emph{eff}_{\bar{a}'}$. A grounded action $\bar{a}'$ is applicable in a state $s$ only if $\emph{pre}_{\bar{a}'} \models s$. Every deterministic grounded action $\bar{a}' \in \bar{\gA}'$ maps each symbolic state $s_i$ to a new state $s_j$.
%  \begin{wrapfigure}[14]{r}{0.30\textwidth}
% % \vspace{-6em}
% % \includegraphics[width=0.35\textwidth]{images/relative.eps}
% \includegraphics[height = 2in]{images/relative.eps}
% \caption{An illustration for computing relative poses}
% \label{fig:relative_poses}
% \end{wrapfigure}

The set $\bar{\gA}$ delineates symbolic lifted actions utilizing the lifted relations $\gV$. Each action $\bar{a} \in \bar{\gA}$ has typed symbolic parameters and is defined using a tuple $\langle \emph{pre}_{\bar{a}}, \emph{eff}_{\bar{a}} \rangle$. Here, $\emph{pre}_{\bar{a}}$ represents a precondition of the action $\bar{a}$ as a conjunction of parameterized relations $\gV$. The effect of the action, $\emph{eff}_a$, is detailed as $\emph{eff}_{\bar{a}} = \langle \emph{add}_{\bar{a}}, \emph{del}_{\bar{a}} \rangle$, which involves adding $\emph{add}_{\bar{a}}$ and removing $\emph{del}_{\bar{a}}$ relations from the state upon execution of action $\bar{a}$. These actions $\bar{a}$ can be instantiated with specific objects to obtain grounded actions $\bar{a'}$, thus producing a grounded precondition $\emph{pre}_{\bar{a'}}$ and effect $\emph{eff}_{\bar{a'}}$. An instantiated action $\bar{a'}$ applies in a state $s'$ only if the precondition $\emph{pre}_{\bar{a'}} \models s'$. Each deterministic grounded action $\bar{a'} \in \bar{\gA'}$ defines a deterministic function $a': s'_i \mapsto s'_j$ that transitions a symbolic state $s'_i$ to another state $s'_j$.

Symbolic plans cannot be executed by a robot. It needs to be converted to a sequence of primitive actions that a robot can execute. Task and motion planning approaches
% ~\citep{srivastava2014combined}
use abstract symbolic models along with \emph{pose generators} for computing a sequence of primitive actions for planning problems. A pose generator defines an inverse abstraction function. Let $\gamma_p$ be a pose generator for a lifted symbolic predicate $p \in \gP$. For a grounded predicate $p'$, a pose generator $\gamma_{p'} = \{ x | x \in \gX \land p'_x = 1  \}$. A pose generator for a grounded state $s'$ is defined as 
$\bigcap_{\forall p'\in s'}\gamma_{p'}$. 

\subsection{Overview of the LAMP}
\label{app:overall}

\begin{figure*}[h]
    \includegraphics[width=\textwidth]{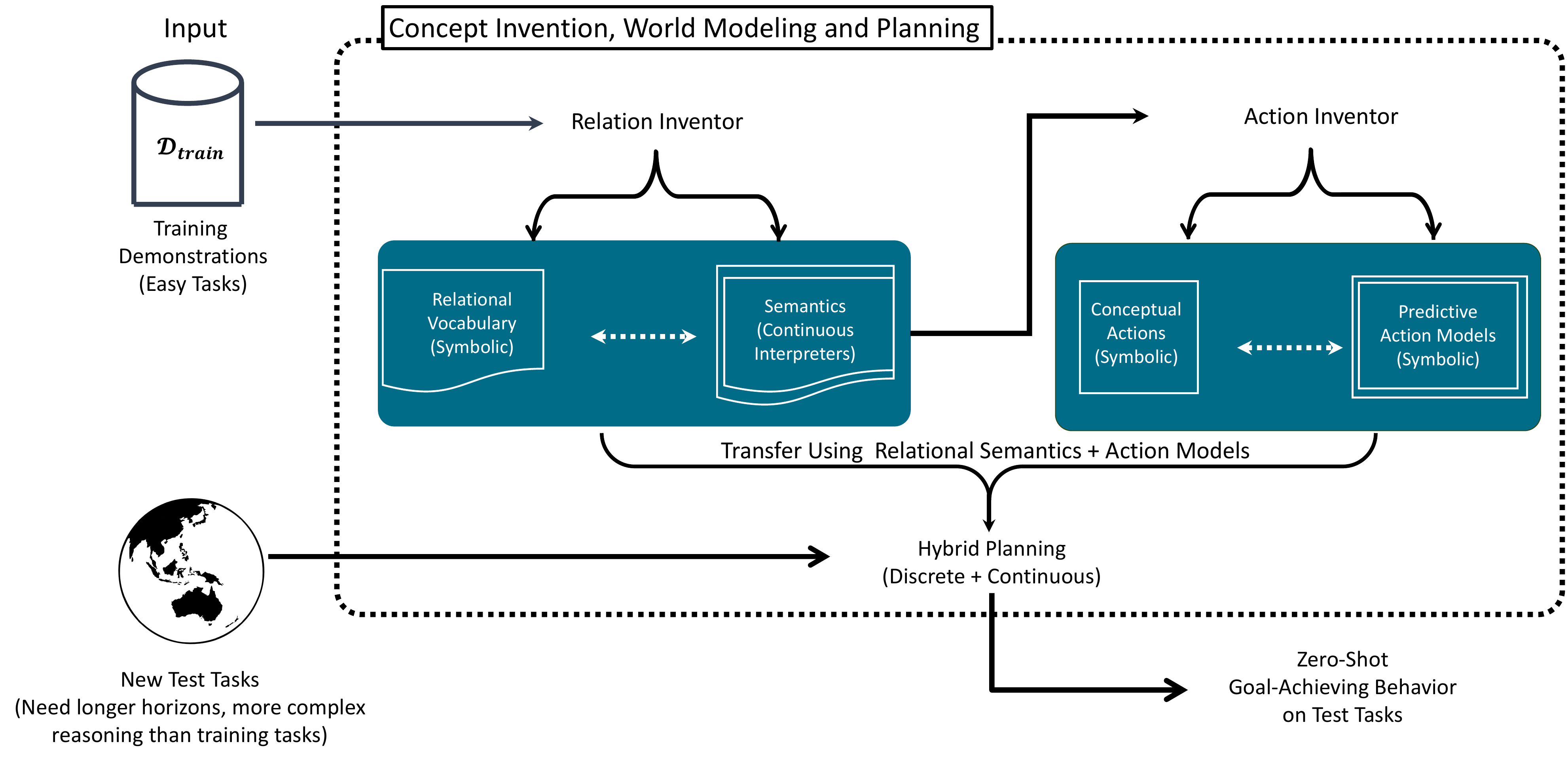}
    \caption{\small Overview of the proposed algorithm. The approach comprises of two core components: $(i)$ ``Relation Inventor'' that uses training demonstrations and generates novel relational vocabularies for robots, and $(ii)$ ``Action Inventor'' that invents high-level symbolic actions. Together, they enable zero-shot goal-achieving behavior by integrating relational semantics, predictive action models, and hybrid planning.}
\end{figure*}

\begin{figure}[h]
\vspace{-1.5em}
    \begin{algorithm}[H]
    {\small 
        \caption{LAMP: Learning Abstract Model for Planning}
        \label{alg:overall}
        \KwIn{A set of demonstrations $\gD_\emph{train}$ for training tasks $T_\emph{train}$, a set of objects $\gO$, a set of types of objects $\gT$, test tasks $T_\emph{test}$}
        \KwOut{Symbolic world model $\gM$}
        
        \tcc{Use Alg. 2 to invent relations}
        $\gV \gets $ Relation\_Inventor($\gD_\emph{train}$); \\ 

        \tcc {Use Alg. 3 to invent actions}
        $\bar{\gA} \gets $ Action\_Inventor($\gD_\emph{train}, \gV$); \\ 

        $\gM = \langle \gV, \bar{\gA} \rangle$; \\ 

        \tcc {Solve new unseen task with task and motion planning (App. F)} 

        $\Pi_{\test} \gets $ task\_and\_motion\_planning($T_\emph{test}, \gM$); \\ 

        \KwRet{$\gM$}; \\    
        }
    \end{algorithm}
\end{figure}

~\newpage
\subsection{Relation Inventor}
\label{app:relation_inventor}

\begin{figure}[h]
    \begin{algorithm}[H]
    {\small 
        \caption{Relation Inventor}
        \label{alg:relation_inventor}
        \KwIn{Training Demonstrations $\gD_\emph{train}$, set of objects $\gO$, set of object types $\gT$}
        \KwOut{Set of relations $\gV$}
    \tcc{Prepare data}
    $\tilde{\gD}_\emph{train} \gets$ $\xi(\gD_\emph{train}, \gO)$ \\ \label{line:augment_data}
    \tcc{Learn relational critical region predictor} 
    $\Psi \gets $ learn\_relational\_critical\_regions\_predictor($\tilde{\gD}_\emph{train}$) \\ \label{line:compute_rcrs}
    \tcc{Identify binary relations}
    $\gV_\emph{bin} \gets $ invent\_binary\_relations($\Psi$) \\ \label{line:invent_binary_relations} 
    \tcc{Identify additional relations}
    $\gV_\emph{add} \gets $ invent\_additional\_relations($\gV_\emph{bin},\Psi)$ \label{line:invent_additional_relations}  \\ 
    $\gV \gets \gV_\emph{bin} \cup \gV_\emph{add} $ \\ 
    \KwRet{$\gV$} 
    }    
    \end{algorithm}
\end{figure}
% ~\newpage
\subsection{Action Inventor}
\label{app:action_inventor}
~
\begin{figure}[h!]
    \begin{algorithm}[H]
    {\small
    \caption{Inventing Symbolic Actions}
    \label{alg:action_inventor}
    \KwIn{Set of demonstrations $\gD_\emph{train}$, learned relations $\gV$}
    \KwOut{Set of lifted actions $\bar{\gA}$}
    $\bar{\gD'}_\emph{train} \gets$ get\_abstract\_demonstrations($\gD_\emph{train}$,$\gV$); \\ \label{line:abstract_demonstrations}
    $\bar{\gD}_\emph{train} \gets$ lift\_demonstrations($\bar{\gD'}_\emph{train}$); \\ \label{line:lifted_demonstrations} 
    changed\_predicates $\gets$ []; \\ 
    \ForEach{$d^k \in \bar{\gD}$} {
        \ForEach{consecutive state $s_i, s_j \in d^k$ }{
            $^{+}C^k_{ij} \gets s_j \setminus s_i$; $^{-}C^k_{ij} \gets s_i \setminus s_j$; \\
             $C^k_{ij} \gets \langle ^{+}C^k_{ij}, ^{-}C^k_{ij} \rangle$; \\ \label{line:changed_predicates}
             changed\_predicates.add($C^{k}_{ij}$); \\ 
        }
    }
    $\gC \gets$ create\_clusters($\bar{\gD}_\emph{train}$, changed\_predicates); \\ \label{line:compute_clusters}
    $\bar{\gA} \gets$ []; \\ 
    \ForEach {$(S_i \rightarrow S_j), C \in \gC$} { 
        $\emph{eff} \gets \langle \emph{add}$ = $^{+}C$, $\emph{del}$ = $^{-}C \rangle$; \\ \label{line:learn_effect}
        $\emph{pre} \gets \cap_{s \in S_i} s$; \\ \label{line:learn_pre}
        $\emph{pre} \gets $prune\_precondition($\emph{pre}$); \\ \label{line:prune_pre}
        $\emph{param} \gets$ extract\_params($S_i \rightarrow S_j$); \\ \label{line:learn_param}
        $\bar{\gA}$.add(create\_action($\emph{param},\emph{pre},\emph{eff}$)); \\ 
    }
    \KwRet{$\bar{\gA}$}
    }
\end{algorithm}
\end{figure}

After identifying a group of high-level actions denoted by $\bar{\gA}$ we employ associative learning along with the training demonstrations $\gD_\emph{train}$ to develop a symbolic representation for each action $\bar{a} \in \bar{\gA}$. The symbolic model of an action is described through its symbolic effects, symbolic preconditions, and parameters. 
% Additionally, our method constructs the symbolic model for each high-level action by utilizing the training demonstrations set $\gD_\emph{train}$ in the following manner.

\paragraph{Learning effects} As noted earlier, in our setting, effect of an action $\bar{a}$ is represented as $\emph{eff}_{\bar{a}} = \langle \emph{add}_{\bar{a}}, \emph{del}_{\bar{a}} \rangle$.  Each cluster $c_i \in \gC$ is generated by clustering transitions the sets of changed relations. These changed relations correspond to added and removed relations as an effect of executing the action induced by the cluster. Therefore, for an action $\bar{a}_i$ induced by the cluster $c_i$ with a set of changed relations $C_i = \langle ^{+}C_i, ^{-}C_i \rangle$,  $\emph{add}_{\bar{a}_i} = ^{+}\!C_i$ and $\emph{del}_{\bar{a}_i} = ^{-}\!C_i$.

\paragraph{Learning preconditions}
To learn the precondition of an action, we take the intersection of all states where the action is applicable. Given a set of relations, this approach generates a maximal precondition that is conservative yet sound~\citep{Wang_1994_Learning,stern_2017_efficient}.   We learn the precondition of an action $\bar{a} \in \bar{\gA}$ corresponding to a cluster $c = \langle S_i \rightarrow S_j, C_{ij} \rangle$ $\emph{pre}_{\bar{a}} = \cap_{s \in S_i} s$. 

Each action can have spurious preconditions corresponding to static relations that do not change when the action is applied, but are still true in all the pre-states. Therefore, we remove relations from the learned precondition that
(i) are not parameterized by any of the objects that are changed by the action and
(ii) are not changed at any point in any of the demonstrations. This removes any predicate from
the precondition that is spurious with respect to the data.

\paragraph{Learning parameters}
Once the precondition and effect of an action are learned, the final step is to learn the parameters of the action that can be replaced with objects in order to ground the action. In this step, the relations in precondition and effect are processed in order. These relations are processed in alphanumeric order and each of their parameters is added to the action's parameter list, if not already added. This process leads to an ordered list of parameters of the action, which can be grounded with compatible objects. 

\subsubsection{Example of Action Model Learning}

Let the set of predicates invented in Sec.~\ref{subsec:relational_vocab} be the following:
\begin{itemize}
    \item \texttt{(table-can0 ?table ?can)}: Can is not on the table.
    \item \texttt{(table-can1  ?table ?can)}: Can is on the table.
    \item \texttt{(can-gripper1 ?can ?gripper)}: Gripper is at grasp pose (not holding/grasping yet).
    \item \texttt{(can-gripper2 ?can ?gripper)}: Gripper has grasped the object.
    \item \texttt{(base-gripper0 ?base ?gripper)}: Robot's base link and robot's gripper link does not have any relation.
    \item \texttt{(base-gripper1 ?base ?gripper)}: Robot's arm is tucked so there is a specific relative pose between the robot's base link and the robot's gripper link.
    \item \texttt{(base-table1 ?base ?table)}: Robot's base link is located in a way such that the robot's arm can reach objects on the table.
\end{itemize}

\begin{figure}[h]
    \centering
    \includegraphics[width=\textwidth]{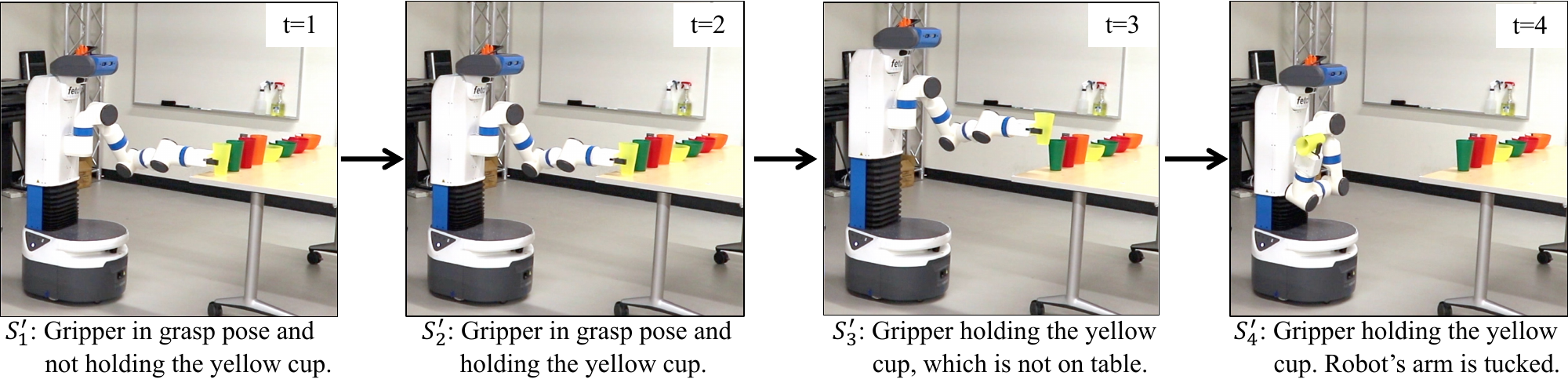}
    \caption{Trajectory $T_1 = \langle S^{'}_1, S^{'}_2, S^{'}_3, S^{'}_4 \rangle$ corresponding to the process of picking up a yellow cup from the table. The state description below each image explains that image in English. These state descriptions are added here for ease of understanding only.}
    \label{fig:trajectory-example-yellow}
\end{figure}

\begin{figure}[h]
    \centering
    \includegraphics[width=\textwidth]{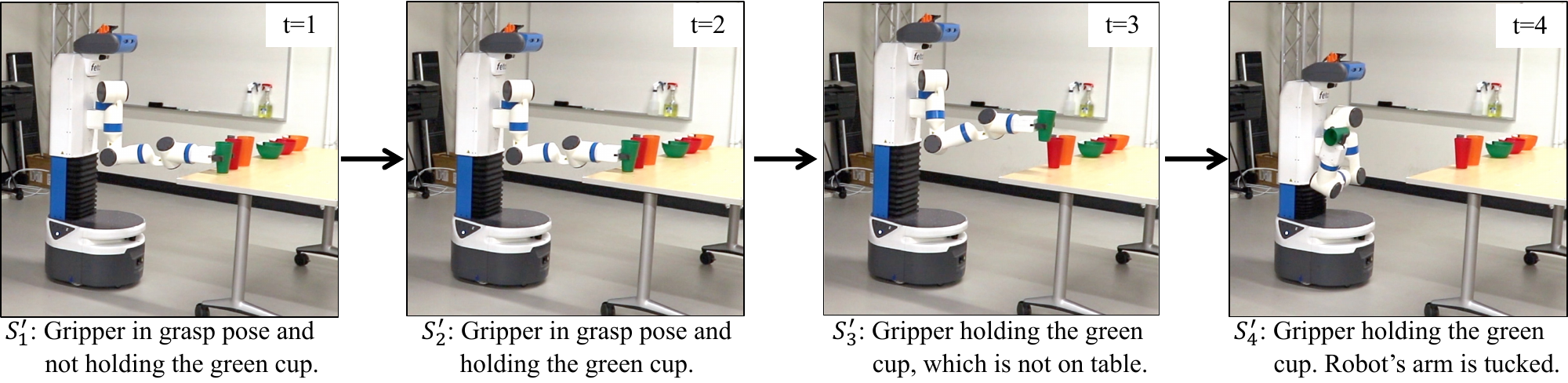}
    \caption{Trajectory $T_2 = \langle S^{'}_1, S^{'}_2, S^{'}_3, S^{'}_4 \rangle$ corresponding to the process of picking up a green cup from the table. The state description below each image explains that image in English. These state descriptions are added here for ease of understanding only.}
    \label{fig:trajectory-example-green}
\end{figure}

Now, consider the two trajectories $T_1$ and $T_2$ as shown in Fig.~\ref{fig:trajectory-example-yellow} and Fig.~\ref{fig:trajectory-example-green}, respectively. Here $T_1$ corresponds to the Fetch robot picking a yellow cup, and $T_2$ corresponds to the robot picking up a green cup (kept at a different location on the table compared to that of the yellow cup). 
Here these two trajectories are expressed in terms of grounded objects. These are converted to a lifted form using line 2 of Alg.~\ref{alg:action_inventor} in terms of the predicates shown earlier.
For $T_1$ and $T_2$ both, the lifted states will be:
\begin{itemize}[leftmargin=*]
    \item $S_1: \{$\texttt{(table-can1 ?table ?can)}, \texttt{(can-gripper1 ?can ?gripper)}, \texttt{(base-gripper0} \texttt{?base ?gripper)}, \texttt{(base-table1} \texttt{?base ?table)}$\}$.
    \item $S_2: \{$\texttt{(table-can1 ?table ?can)}, \texttt{(can-gripper2 ?can ?gripper)}, \texttt{(base-gripper0} \texttt{?base ?gripper)}, \texttt{(base-table1} \texttt{?base ?table)}$\}$.
    \item $S_3: \{$\texttt{(table-can0 ?table ?can)}, \texttt{(can-gripper2 ?can ?gripper)}, \texttt{(base-gripper0} \texttt{?base ?gripper)}, \texttt{(base-table1} \texttt{?base ?table)}$\}$.
    \item $S_4: \{$\texttt{(table-can0 ?table ?can)}, \texttt{(can-gripper2 ?can ?gripper)}, \texttt{(base-gripper1} \texttt{?base ?gripper)}, \texttt{(base-table1} \texttt{?base ?table)}$\}$.
\end{itemize}

Note that we only show partial states here for brevity. The actual states will also have predicates like \texttt{(table-can1 ?table ?can2)}, \texttt{(table-can1 ?table ?can3)}, \texttt{(table-can1 ?table ?can4)}, \texttt{(table-can1 ?table ?bowl1)}, \texttt{(table-can1 ?table ?bowl2)}, \texttt{(table-can1 ?table ?bowl3)}, etc. corresponding to other objects kept on the table.

\paragraph{Learning effects} Alg.~\ref{alg:action_inventor} creates the following three clusters (lines 4-9) based on these states.
\begin{itemize}[leftmargin=*]
    \item $C_{12} = \langle ^{+}C_{12} = \{$\texttt{(can-gripper2 ?can ?gripper)}$\},^{-}C_{12} = \{$\texttt{(can-gripper1 ?can ?gripper)}$\} \rangle$.
    \item $C_{23} = \langle ^{+}C_{23} = \{$\texttt{(table-can0 ?table ?can)}$\}, ^{-}C_{23} = \{$\texttt{(table-can1 ?table ?can)}$\}\rangle$.
    \item $C_{34} = \langle ^{+}C_{34} = \{$\texttt{(base-gripper1 ?base ?gripper)}$\}, ^{-}C_{34} = \{$\texttt{(base-gripper0 ?base ?gripper)}$\}\rangle$.
\end{itemize}

\paragraph{Learning preconditions} Learning preconditions involve taking intersection of states in which all the actions in the same cluster were executed. Here $S_1$ to $S_3$ mentioned below will remain the same for the three clusters. E.g., precondition of $C_{12} = \{$\texttt{(table-can1 ?table ?can)}, \texttt{(can-gripper1 ?can ?gripper)}, \texttt{(base-gripper0} \texttt{?base ?gripper)}, \texttt{(base-table1} \texttt{?base ?table)}$\}$. Alg.~\ref{alg:action_inventor} will prune out \texttt{(base-table1} \texttt{?base ?table)} from the precondition as (i) it is unchanged between $S_1$ and $S_2$, and (ii) none of its parameters (\texttt{?base} and \texttt{?table}) are part of any other predicate that is changed. Using this, the precondition for each action will be:
\begin{itemize}[leftmargin=*]
    \item $\emph{pre}(C_{12}) =  \{$\texttt{(table-can1 ?table ?can)}, \texttt{(can-gripper1 ?can ?gripper)}, \texttt{(base-gripper0} \texttt{?base ?gripper)}$\}$.
    \item $\emph{pre}(C_{23}) =  \{$\texttt{(table-can1 ?table ?can)}, \texttt{(can-gripper2 ?can ?gripper)}, \texttt{(base-gripper0} \texttt{?base ?gripper)}, \texttt{(base-table1} \texttt{?base ?table)}$\}$.
    \item $\emph{pre}(C_{34}) =  \{$\texttt{(table-can0 ?table ?can)}, \texttt{(can-gripper2 ?can ?gripper)}, \texttt{(base-gripper0} \texttt{?base ?gripper)}, \texttt{(base-table1} \texttt{?base ?table)}$\}$.
\end{itemize}

\paragraph{Learning parameters} Learning parameters from an action's precondition and effect is straightforward. All the unique parameters in predicates in the precondition and effect are added to the parameter list of an action representing a cluster. Using this notion, the parameters for the three clusters will be the following:

\begin{itemize}
    \item $\emph{param}(C_{12}) = $\texttt{(?table ?can ?gripper ?base)}.
    \item $\emph{param}(C_{23}) = $\texttt{(?table ?can ?gripper ?base)}.
    \item $\emph{param}(C_{34}) = $\texttt{(?table ?can ?gripper ?base)}.
\end{itemize}

\subsection{Related Work}
\label{app:related_work}
The presented approach directly relates to various concepts in task and motion planning, model learning, and abstraction learning. However, to the best of our knowledge, this is the first work that automatically invents generalizable symbolic predicates and high-level actions simultaneously using a set of low-level trajectories. 

\paragraph{Task and motion planning} Task and motion planning approaches~\citep{srivastava2014combined,dantam2018incremental,garrett2020pddlstream,shah2020anytime} develop approaches for autonomously solving long-horizon robot planning problems. These approaches are complementary to the presented approach as they focus on using provided abstractions for efficiently solving the robot planning problems. \citet{shah2022using,shah2024hierarchical} learn state and action abstractions for long-horizon motion planning problems. Orthogonal research~\citep{mishra2023generative,cheng2023nod,fang2023dimsam} learn implicit abstractions (action interpreters or abstract actions) for TAMP in the form of generative models. However, these approaches do not learn generalizable relational representations as well as complex high-level relations and actions which is the focus of our work.  

% Several approaches have focused on learning symbolic predicates for low-level continuous state spaces. 
\paragraph{Learning symbolic abstractions} Several approaches invent symbolic vocabularies given a set of high-level actions (or skills) ~\citep{konidaris2014constructing,ugur2015bottom,konidaris2015symbol,andersen2017active,konidaris2018skills,bonet2019learning,james2020learning}. \citet{ahmetoglu2022deepsym,asai2022classical,liang2023learning} learn symbolic predicates in the form of latent spaces of deep neural networks and use them for high-level symbolic planning. However, these approaches assume high-level actions to be provided as input. On the other hand, the approach presented in this paper automatically learns high-level actions along with symbolic predicates.

Numerous approaches~\citep{Yang2007,Cresswell09,zhuo2013action,aineto2019learning,verma_21_asking} have focused on learning preconditions and effects for high-level actions, i.e., action model. A few  approaches~\citep{certicky_2014_real,lamanna_2021_online} have also focused on continually learning action models while collecting experience in the environment. \citet{Bryce2016} and \citet{nayyar2022differential} focus on updating a known model using inconsistent observations. However, these approaches require a set of symbolic predicates and/or high-level action signatures as input whereas our approach automatically invents these predicates and actions.  Several approaches~\citep{silver2021learning, verma2022discovering,chitnis2022learning, silver2022learning, kumar2023learning, silver2023PredicateInvention} have been able to automatically invent high-level actions that are induced by state abstraction akin to the presented approach. However, unlike our approach, these approaches do not automatically learn symbolic predicates and/or low-level samplers and require them as input.  

\paragraph{Behavior Cloning for Robotics}
Behavior cloning (BC) has been widely explored in robotics as a method for learning control policies from expert demonstrations. Early work by \citet{pomerleau1988alvinn} introduced the concept of imitation learning through supervised learning, where a model maps sensor inputs to control actions. Recent years have seen renewed interest in behavior cloning approaches due to the rise in applicability of deep neural networks~\citep{levine2016end,brohan2022rt,fu2024mobile,zhaoaloha,black2024pi_0}. These appraoches have shown remarkable advancements in dexterous manipulation capabilities (e.g., tying shoelaces and chopping fruits) and ability to manipulate deformable objects (e.g., folding laundry). However, these approaches are only limited to tackling problems in their training demonstrations as well as often require huge amounts of training demonstrations (in the order of 1000s of demonstrations per task), reducing the applicability of such approaches in real-world settings where the distribution of tasks is not known a priori.  

\paragraph{LLMs for robot planning}
Recent years have also seen significantly increased interest in using foundational models such as LLM (large language model), VLM (visual language model), and transformers for robot planning and control owing to their success in other fields such as NLP, text generation, and vision. Several approaches \citep{brohan2022rt, goyal2023rvt,shridhar2023perceiver,padalkar2023open} use transformer architecture for learning reactive policies for short-horizon robot control problems. Problems tackled by these approaches are analogous to individual actions learned by our approach. 

Several directions of research explore the use of LLMs for utilizing LLMs as high-level planners to generate sequences comprising of high-level, expert crafted actions \citep{yu2023language, liang2023code,huang2022language, rana2023sayplan, lin2023text2motion, huang2023inner, brohan2023can}.  These  methods make progress on the problem of near-natural language communication with robots and are complementary to the proposed work. However, there is a strong evidence against the soundness of LLMs as planners. \citet{valmeekam2023planning} show that LLMs are only $\sim 36\%$ accurate as planners even in simple block stacking settings not involving more than $5$ object.

\subsection{Task and Motion Planning with Learned World Models}
\label{app:tamp}

Task and motion planning~\citep{srivastava2014combined,garrett2020pddlstream,shah2020anytime} combines symbolic world representations and action interpreters to develop an interleaved planning approach. This approach seeks a valid sequence of high-level actions each accompanied by legitimate primitive action refinement. We apply a deterministic form of STAMP~\citep{shah2020anytime} for task and motion planning relying on learned world representations. STAMP utilizes the acquired relational vocabulary along with high-level robotic actions to generate a high-level action sequence, subsequently refining these actions using learned interpreters into executable primitive actions. 

One of the major challenges in TAMP with automatically learned world models is the overly pessimistic or conservative nature of these models, often leading the planner to a failed search even in the presence of potential plans. To address this, we incorporate predicate abstractions~\citep{graf1997construction} to systematically create a relaxed high-level planning problem that STAMP can work to refine. However, the use of predicate abstractions may result in inaccurate world models, thus permitting high-level plans lacking viable refinements. To counteract this, we enhance the STAMP high-level planner to a top-$k$~\citep{katz-et-al-icaps2018} version, generating multiple high-level plans. We then progressively refine these $k$ plans until we achieve a plan supporting valid primitive action refinements for each high-level robot activity.

\subsection{Code and Data}
Code and data are available on the project webpage at \url{https://aair-lab.github.io/r2l-lamp}. 

\subsection{Learned Word Models}
\subsubsection{Delivering Items in a Cafe}

\begin{Verbatim}[breaklines=true]
(define (domain CafeWorld)
(:requirements :strips :typing :equality :conditional-effects :existential-preconditions :universal-preconditions)
(:types 
	freight
	can
	gripper
	surface
)

(:constants 
)

(:predicates 
	(freight_surface_0 ?x - freight ?y - surface)
	(freight_surface_1 ?x - freight ?y - surface)
	(gripper_can_0 ?x - gripper ?y - can)
	(gripper_can_1 ?x - gripper ?y - can)
	(gripper_can_2 ?x - gripper ?y - can)
	(freight_gripper_0 ?x - freight ?y - gripper)
	(freight_gripper_1 ?x - freight ?y - gripper)
	(can_surface_0 ?x - can ?y - surface)
	(can_surface_1 ?x - can ?y - surface)
	(freight_can_0 ?x - freight ?y - can)
	(freight_can_1 ?x - freight ?y - can)
	(clear3_gripper_can_1 ?x - gripper)
	(clear3_freight_can_1 ?x - freight)
	(clear3_freight_gripper_1 ?x - freight)
	(clear3_freight_surface_1 ?x - freight)
	(clear3_can_surface_1 ?x - can)
	(clear3_gripper_can_2 ?x - gripper)
)

(:action a1 
:parameters ( ?can_p1 - can  ?freight_p1 - freight  ?surface_extra_p1 - surface  ?gripper_p1 - gripper )
:precondition (and 
	(can_surface_1 ?can_p1 ?surface_extra_p1)
	(freight_can_0 ?freight_p1 ?can_p1)
	(freight_gripper_0 ?freight_p1 ?gripper_p1)
	(freight_surface_1 ?freight_p1 ?surface_extra_p1)
	(gripper_can_2 ?gripper_p1 ?can_p1)
	(clear3_gripper_can_1 ?gripper_p1) 
) 
:effect (and 
 	(gripper_can_1 ?gripper_p1 ?can_p1) 
	(not (gripper_can_0 ?gripper_p1 ?can_p1))
	(not (gripper_can_2 ?gripper_p1 ?can_p1))
	(clear3_gripper_can_2 ?gripper_p1) 
	(not (clear3_gripper_can_1 ?gripper_p1)) 
 ) 
)

(:action a2 
:parameters ( ?surface_extra_p4 - surface  ?can_p1 - can  ?gripper_p1 - gripper  ?surface_extra_p1 - surface  ?freight_p1 - freight )
:precondition (and 
	(not (= ?surface_extra_p4 ?surface_extra_p1))
	(can_surface_0 ?can_p1 ?surface_extra_p4)
	(freight_can_0 ?freight_p1 ?can_p1)
	(freight_gripper_0 ?freight_p1 ?gripper_p1)
	(freight_surface_0 ?freight_p1 ?surface_extra_p4)
	(freight_surface_1 ?freight_p1 ?surface_extra_p1)
	(gripper_can_2 ?gripper_p1 ?can_p1)
	(clear3_freight_can_1 ?freight_p1) 
	(clear3_freight_gripper_1 ?freight_p1) 
) 
:effect (and 
 	(freight_can_1 ?freight_p1 ?can_p1) 
	(freight_gripper_1 ?freight_p1 ?gripper_p1) 
	(not (freight_can_0 ?freight_p1 ?can_p1))
	(not (freight_gripper_0 ?freight_p1 ?gripper_p1))
	(not (clear3_freight_can_1 ?freight_p1)) 
	(not (clear3_freight_gripper_1 ?freight_p1)) 
 ) 
)

(:action a3 
:parameters ( ?can_p1 - can  ?gripper_p1 - gripper  ?surface_extra_p1 - surface  ?freight_p1 - freight )
:precondition (and 
	(can_surface_1 ?can_p1 ?surface_extra_p1)
	(freight_can_0 ?freight_p1 ?can_p1)
	(freight_gripper_0 ?freight_p1 ?gripper_p1)
	(freight_surface_1 ?freight_p1 ?surface_extra_p1)
	(gripper_can_1 ?gripper_p1 ?can_p1)
	(clear3_gripper_can_2 ?gripper_p1) 
) 
:effect (and 
 	(gripper_can_0 ?gripper_p1 ?can_p1) 
	(not (gripper_can_1 ?gripper_p1 ?can_p1))
	(not (gripper_can_2 ?gripper_p1 ?can_p1))
	(clear3_gripper_can_1 ?gripper_p1) 
 ) 
)

(:action a4 
:parameters ( ?can_p1 - can  ?gripper_p1 - gripper  ?surface_extra_p1 - surface  ?freight_p1 - freight )
:precondition (and 
	(freight_can_1 ?freight_p1 ?can_p1)
	(freight_gripper_1 ?freight_p1 ?gripper_p1)
	(freight_surface_1 ?freight_p1 ?surface_extra_p1)
	(gripper_can_2 ?gripper_p1 ?can_p1)
) 
:effect (and 
 	(freight_can_0 ?freight_p1 ?can_p1) 
	(freight_gripper_0 ?freight_p1 ?gripper_p1) 
	(not (freight_can_1 ?freight_p1 ?can_p1))
	(not (freight_gripper_1 ?freight_p1 ?gripper_p1))
	(clear3_freight_can_1 ?freight_p1) 
	(clear3_freight_gripper_1 ?freight_p1) 
 ) 
)

(:action a5 
:parameters ( ?gripper_extra_p1 - gripper  ?can_p1 - can  ?freight_extra_p1 - freight  ?surface_p1 - surface )
:precondition (and 
	(can_surface_0 ?can_p1 ?surface_p1)
	(freight_can_0 ?freight_extra_p1 ?can_p1)
	(freight_gripper_0 ?freight_extra_p1 ?gripper_extra_p1)
	(freight_surface_1 ?freight_extra_p1 ?surface_p1)
	(gripper_can_2 ?gripper_extra_p1 ?can_p1)
	(clear3_can_surface_1 ?can_p1) 
) 
:effect (and 
 	(can_surface_1 ?can_p1 ?surface_p1) 
	(not (can_surface_0 ?can_p1 ?surface_p1))
	(not (clear3_can_surface_1 ?can_p1)) 
 ) 
)

(:action a6 
:parameters ( ?gripper_p1 - gripper  ?surface_extra_p1 - surface  ?freight_p1 - freight )
:precondition (and 
	(freight_gripper_1 ?freight_p1 ?gripper_p1)
	(freight_surface_1 ?freight_p1 ?surface_extra_p1)
) 
:effect (and 
 	(freight_gripper_0 ?freight_p1 ?gripper_p1) 
	(not (freight_gripper_1 ?freight_p1 ?gripper_p1))
	(clear3_freight_gripper_1 ?freight_p1) 
 ) 
)

(:action a7 
:parameters ( ?can_p1 - can  ?gripper_p1 - gripper  ?surface_extra_p1 - surface  ?freight_p1 - freight )
:precondition (and 
	(can_surface_1 ?can_p1 ?surface_extra_p1)
	(freight_can_0 ?freight_p1 ?can_p1)
	(freight_gripper_0 ?freight_p1 ?gripper_p1)
	(freight_surface_1 ?freight_p1 ?surface_extra_p1)
	(gripper_can_1 ?gripper_p1 ?can_p1)
	(clear3_gripper_can_2 ?gripper_p1) 
) 
:effect (and 
 	(gripper_can_2 ?gripper_p1 ?can_p1) 
	(not (gripper_can_0 ?gripper_p1 ?can_p1))
	(not (gripper_can_1 ?gripper_p1 ?can_p1))
	(clear3_gripper_can_1 ?gripper_p1) 
	(not (clear3_gripper_can_2 ?gripper_p1)) 
 ) 
)

(:action a8 
:parameters ( ?gripper_p1 - gripper  ?surface_extra_p1 - surface  ?freight_p1 - freight )
:precondition (and 
	(freight_gripper_0 ?freight_p1 ?gripper_p1)
	(freight_surface_1 ?freight_p1 ?surface_extra_p1)
	(clear3_freight_gripper_1 ?freight_p1) 
) 
:effect (and 
 	(freight_gripper_1 ?freight_p1 ?gripper_p1) 
	(not (freight_gripper_0 ?freight_p1 ?gripper_p1))
	(not (clear3_freight_gripper_1 ?freight_p1)) 
 ) 
)

(:action a9 
:parameters ( ?can_p1 - can  ?freight_p1 - freight  ?surface_extra_p1 - surface  ?gripper_p1 - gripper )
:precondition (and 
	(can_surface_1 ?can_p1 ?surface_extra_p1)
	(freight_can_0 ?freight_p1 ?can_p1)
	(freight_gripper_0 ?freight_p1 ?gripper_p1)
	(freight_surface_1 ?freight_p1 ?surface_extra_p1)
	(gripper_can_0 ?gripper_p1 ?can_p1)
	(clear3_gripper_can_1 ?gripper_p1) 
	(clear3_gripper_can_2 ?gripper_p1) 
) 
:effect (and 
 	(gripper_can_1 ?gripper_p1 ?can_p1) 
	(not (gripper_can_0 ?gripper_p1 ?can_p1))
	(not (gripper_can_2 ?gripper_p1 ?can_p1))
	(not (clear3_gripper_can_1 ?gripper_p1)) 
 ) 
)

(:action a10 
:parameters ( ?gripper_p1 - gripper  ?surface_p1 - surface  ?freight_p1 - freight )
:precondition (and 
	(freight_gripper_1 ?freight_p1 ?gripper_p1)
	(freight_surface_0 ?freight_p1 ?surface_p1)
	(clear3_freight_surface_1 ?freight_p1) 
) 
:effect (and 
 	(freight_surface_1 ?freight_p1 ?surface_p1) 
	(not (freight_surface_0 ?freight_p1 ?surface_p1))
	(not (clear3_freight_surface_1 ?freight_p1)) 
 ) 
)

(:action a11 
:parameters ( ?gripper_extra_p1 - gripper  ?can_p1 - can  ?freight_extra_p1 - freight  ?surface_p1 - surface )
:precondition (and 
	(can_surface_1 ?can_p1 ?surface_p1)
	(freight_can_0 ?freight_extra_p1 ?can_p1)
	(freight_gripper_0 ?freight_extra_p1 ?gripper_extra_p1)
	(freight_surface_1 ?freight_extra_p1 ?surface_p1)
	(gripper_can_2 ?gripper_extra_p1 ?can_p1)
) 
:effect (and 
 	(can_surface_0 ?can_p1 ?surface_p1) 
	(not (can_surface_1 ?can_p1 ?surface_p1))
	(clear3_can_surface_1 ?can_p1) 
 ) 
)

(:action a12 
:parameters ( ?gripper_p1 - gripper  ?surface_p1 - surface  ?freight_p1 - freight )
:precondition (and 
	(freight_gripper_1 ?freight_p1 ?gripper_p1)
	(freight_surface_1 ?freight_p1 ?surface_p1)
) 
:effect (and 
 	(freight_surface_0 ?freight_p1 ?surface_p1) 
	(not (freight_surface_1 ?freight_p1 ?surface_p1))
	(clear3_freight_surface_1 ?freight_p1) 
 ) 
))
\end{Verbatim}

\subsubsection{Setting Up a Dinner Table}



\subsubsection{Building Structures with Keva Planks}

%

\subsubsection{Building Structures with Jenga Planks}

\begin{Verbatim}[breaklines=true]
(define (domain Jenga)
(:requirements :strips :typing :equality 
:conditional-effects :existential-preconditions :universal-preconditions)
(:types 
	loc
	jenga
	gripper
)

(:constants 
	   loc_jengaTarget_Const - loc
 )

(:predicates 
    	(loc_jenga_0 ?x - loc ?y - jenga)
    	(loc_jenga_1 ?x - loc ?y - jenga)
    	(loc_jenga_2 ?x - loc ?y - jenga)
    	(jenga_jenga_0 ?x - jenga ?y - jenga)
    	(jenga_jenga_1 ?x - jenga ?y - jenga)
    	(jenga_jenga_2 ?x - jenga ?y - jenga)
    	(jenga_jenga_3 ?x - jenga ?y - jenga)
    	(jenga_jenga_4 ?x - jenga ?y - jenga)
    	(jenga_jenga_5 ?x - jenga ?y - jenga)
    	(jenga_jenga_6 ?x - jenga ?y - jenga)
    	(jenga_jenga_7 ?x - jenga ?y - jenga)
    	(jenga_jenga_8 ?x - jenga ?y - jenga)
    	(gripper_jenga_0 ?x - gripper ?y - jenga)
    	(gripper_jenga_1 ?x - gripper ?y - jenga)
    	(gripper_jenga_2 ?x - gripper ?y - jenga)
    	(clear3_gripper_jenga_1 ?x - gripper)
    	(clear3_jenga_jenga_6 ?x - jenga)
    	(clear3_jenga_jenga_2 ?x - jenga)
    	(clear3_jenga_jenga_3 ?x - jenga)
    	(clear3_jenga_jenga_1 ?x - jenga)
    	(clear3_jenga_jenga_8 ?x - jenga)
    	(clear3_gripper_jenga_2 ?x - gripper)
    	(clear3_jenga_jenga_4 ?x - jenga)
    	(clear3_jenga_jenga_7 ?x - jenga)
    	(clear3_jenga_jenga_5 ?x - jenga)
)

(:action a1 
    :parameters ( ?jenga_p1 - jenga  ?gripper_extra_p1 - gripper )
    :precondition (and 
        	(loc_jenga_0 loc_jengaTarget_Const ?jenga_p1)
        	(gripper_jenga_1 ?gripper_extra_p1 ?jenga_p1)
    ) 
    :effect (and 
         	(loc_jenga_1 loc_jengaTarget_Const ?jenga_p1) 
        	(not (loc_jenga_0 loc_jengaTarget_Const ?jenga_p1))
        	(not (loc_jenga_2 loc_jengaTarget_Const ?jenga_p1))
    ) 
)

(:action a2 
    :parameters ( ?jenga_p1 - jenga  ?gripper_p1 - gripper )
    :precondition (and 
        	(gripper_jenga_0 ?gripper_p1 ?jenga_p1)
        	(clear3_gripper_jenga_1 ?gripper_p1) 
        	(clear3_gripper_jenga_2 ?gripper_p1) 
    ) 
    :effect (and 
         	(gripper_jenga_2 ?gripper_p1 ?jenga_p1) 
        	(not (gripper_jenga_0 ?gripper_p1 ?jenga_p1))
        	(not (gripper_jenga_1 ?gripper_p1 ?jenga_p1))
        	(not (clear3_gripper_jenga_2 ?gripper_p1)) 
    ) 
)

(:action a3 
    :parameters ( ?jenga_p2 - jenga  ?jenga_p3 - jenga  ?jenga_p1 - jenga 
        ?gripper_extra_p1 - gripper )
    :precondition (and 
        	(not (= ?jenga_p2 ?jenga_p3))
        	(not (= ?jenga_p2 ?jenga_p1))
        	(not (= ?jenga_p3 ?jenga_p1))
        	(loc_jenga_0 loc_jengaTarget_Const ?jenga_p2)
        	(loc_jenga_1 loc_jengaTarget_Const ?jenga_p1)
        	(loc_jenga_1 loc_jengaTarget_Const ?jenga_p3)
        	(gripper_jenga_0 ?gripper_extra_p1 ?jenga_p1)
        	(gripper_jenga_0 ?gripper_extra_p1 ?jenga_p3)
        	(gripper_jenga_1 ?gripper_extra_p1 ?jenga_p2)
        	(jenga_jenga_0 ?jenga_p1 ?jenga_p2)
        	(jenga_jenga_0 ?jenga_p3 ?jenga_p2)
        	(clear3_jenga_jenga_1 ?jenga_p1) 
        	(clear3_jenga_jenga_1 ?jenga_p2) 
        	(clear3_jenga_jenga_1 ?jenga_p3) 
        	(clear3_jenga_jenga_3 ?jenga_p1) 
        	(clear3_jenga_jenga_3 ?jenga_p2) 
        	(clear3_jenga_jenga_3 ?jenga_p3) 
    ) 
    :effect (and 
         	(loc_jenga_1 loc_jengaTarget_Const ?jenga_p2) 
        	(jenga_jenga_1 ?jenga_p1 ?jenga_p2) 
        	(jenga_jenga_3 ?jenga_p3 ?jenga_p2) 
        	(not (loc_jenga_0 loc_jengaTarget_Const ?jenga_p2))
        	(not (loc_jenga_2 loc_jengaTarget_Const ?jenga_p2))
        	(not (jenga_jenga_0 ?jenga_p1 ?jenga_p2))
        	(not (jenga_jenga_0 ?jenga_p3 ?jenga_p2))
        	(not (jenga_jenga_1 ?jenga_p3 ?jenga_p2))
        	(not (jenga_jenga_2 ?jenga_p1 ?jenga_p2))
        	(not (jenga_jenga_2 ?jenga_p3 ?jenga_p2))
        	(not (jenga_jenga_3 ?jenga_p1 ?jenga_p2))
        	(not (jenga_jenga_4 ?jenga_p1 ?jenga_p2))
        	(not (jenga_jenga_4 ?jenga_p3 ?jenga_p2))
        	(not (jenga_jenga_5 ?jenga_p1 ?jenga_p2))
        	(not (jenga_jenga_5 ?jenga_p3 ?jenga_p2))
        	(not (jenga_jenga_6 ?jenga_p1 ?jenga_p2))
        	(not (jenga_jenga_6 ?jenga_p3 ?jenga_p2))
        	(not (jenga_jenga_7 ?jenga_p1 ?jenga_p2))
        	(not (jenga_jenga_7 ?jenga_p3 ?jenga_p2))
        	(not (jenga_jenga_8 ?jenga_p1 ?jenga_p2))
        	(not (jenga_jenga_8 ?jenga_p3 ?jenga_p2))
        	(not (clear3_jenga_jenga_1 ?jenga_p1)) 
        	(not (clear3_jenga_jenga_3 ?jenga_p3)) 
     ) 
)

(:action a4 
    :parameters ( ?jenga_p2 - jenga  ?jenga_p3 - jenga  ?jenga_p1 - jenga  
        ?gripper_extra_p1 - gripper )
    :precondition (and 
        	(not (= ?jenga_p2 ?jenga_p3))
        	(not (= ?jenga_p2 ?jenga_p1))
        	(not (= ?jenga_p3 ?jenga_p1))
        	(loc_jenga_0 loc_jengaTarget_Const ?jenga_p1)
        	(loc_jenga_1 loc_jengaTarget_Const ?jenga_p2)
        	(loc_jenga_1 loc_jengaTarget_Const ?jenga_p3)
        	(gripper_jenga_0 ?gripper_extra_p1 ?jenga_p2)
        	(gripper_jenga_0 ?gripper_extra_p1 ?jenga_p3)
        	(gripper_jenga_1 ?gripper_extra_p1 ?jenga_p1)
        	(jenga_jenga_0 ?jenga_p2 ?jenga_p1)
        	(jenga_jenga_0 ?jenga_p3 ?jenga_p1)
        	(clear3_jenga_jenga_2 ?jenga_p1) 
        	(clear3_jenga_jenga_2 ?jenga_p2) 
        	(clear3_jenga_jenga_2 ?jenga_p3) 
        	(clear3_jenga_jenga_8 ?jenga_p1) 
        	(clear3_jenga_jenga_8 ?jenga_p2) 
        	(clear3_jenga_jenga_8 ?jenga_p3) 
    ) 
    :effect (and 
         	(loc_jenga_1 loc_jengaTarget_Const ?jenga_p1) 
        	(jenga_jenga_2 ?jenga_p2 ?jenga_p1) 
        	(jenga_jenga_8 ?jenga_p3 ?jenga_p1) 
        	(not (loc_jenga_0 loc_jengaTarget_Const ?jenga_p1))
        	(not (loc_jenga_2 loc_jengaTarget_Const ?jenga_p1))
        	(not (jenga_jenga_0 ?jenga_p2 ?jenga_p1))
        	(not (jenga_jenga_0 ?jenga_p3 ?jenga_p1))
        	(not (jenga_jenga_1 ?jenga_p2 ?jenga_p1))
        	(not (jenga_jenga_1 ?jenga_p3 ?jenga_p1))
        	(not (jenga_jenga_2 ?jenga_p3 ?jenga_p1))
        	(not (jenga_jenga_3 ?jenga_p2 ?jenga_p1))
        	(not (jenga_jenga_3 ?jenga_p3 ?jenga_p1))
        	(not (jenga_jenga_4 ?jenga_p2 ?jenga_p1))
        	(not (jenga_jenga_4 ?jenga_p3 ?jenga_p1))
        	(not (jenga_jenga_5 ?jenga_p2 ?jenga_p1))
        	(not (jenga_jenga_5 ?jenga_p3 ?jenga_p1))
        	(not (jenga_jenga_6 ?jenga_p2 ?jenga_p1))
        	(not (jenga_jenga_6 ?jenga_p3 ?jenga_p1))
        	(not (jenga_jenga_7 ?jenga_p2 ?jenga_p1))
        	(not (jenga_jenga_7 ?jenga_p3 ?jenga_p1))
        	(not (jenga_jenga_8 ?jenga_p2 ?jenga_p1))
        	(not (clear3_jenga_jenga_2 ?jenga_p2)) 
        	(not (clear3_jenga_jenga_8 ?jenga_p3)) 
     ) 
)

(:action a5 
    :parameters ( ?jenga_p1 - jenga  ?gripper_p1 - gripper )
    :precondition (and 
        	(loc_jenga_1 loc_jengaTarget_Const ?jenga_p1)
        	(gripper_jenga_1 ?gripper_p1 ?jenga_p1)
        	(clear3_gripper_jenga_2 ?gripper_p1) 
    ) 
    :effect (and 
         	(gripper_jenga_2 ?gripper_p1 ?jenga_p1) 
        	(not (gripper_jenga_0 ?gripper_p1 ?jenga_p1))
        	(not (gripper_jenga_1 ?gripper_p1 ?jenga_p1))
        	(clear3_gripper_jenga_1 ?gripper_p1) 
        	(not (clear3_gripper_jenga_2 ?gripper_p1)) 
    ) 
)

(:action a6 
    :parameters ( ?jenga_p1 - jenga  ?gripper_p1 - gripper )
    :precondition (and 
        	(loc_jenga_1 loc_jengaTarget_Const ?jenga_p1)
        	(gripper_jenga_2 ?gripper_p1 ?jenga_p1)
        	(clear3_gripper_jenga_1 ?gripper_p1) 
    ) 
    :effect (and 
         	(gripper_jenga_0 ?gripper_p1 ?jenga_p1) 
        	(not (gripper_jenga_1 ?gripper_p1 ?jenga_p1))
        	(not (gripper_jenga_2 ?gripper_p1 ?jenga_p1))
        	(clear3_gripper_jenga_2 ?gripper_p1) 
    ) 
)

(:action a7 
    :parameters ( ?jenga_p1 - jenga  ?gripper_p1 - gripper )
    :precondition (and 
        	(gripper_jenga_2 ?gripper_p1 ?jenga_p1)
        	(clear3_gripper_jenga_1 ?gripper_p1) 
    ) 
    :effect (and 
         	(gripper_jenga_1 ?gripper_p1 ?jenga_p1) 
        	(not (gripper_jenga_0 ?gripper_p1 ?jenga_p1))
        	(not (gripper_jenga_2 ?gripper_p1 ?jenga_p1))
        	(clear3_gripper_jenga_2 ?gripper_p1) 
        	(not (clear3_gripper_jenga_1 ?gripper_p1)) 
    ) 
)

(:action a8 
    :parameters ( ?jenga_p2 - jenga  ?jenga_p3 - jenga  ?jenga_p1 - jenga  
        ?gripper_extra_p1 - gripper )
    :precondition (and 
        	(not (= ?jenga_p2 ?jenga_p3))
        	(not (= ?jenga_p2 ?jenga_p1))
        	(not (= ?jenga_p3 ?jenga_p1))
        	(loc_jenga_0 loc_jengaTarget_Const ?jenga_p2)
        	(loc_jenga_1 loc_jengaTarget_Const ?jenga_p1)
        	(loc_jenga_1 loc_jengaTarget_Const ?jenga_p3)
        	(gripper_jenga_0 ?gripper_extra_p1 ?jenga_p1)
        	(gripper_jenga_0 ?gripper_extra_p1 ?jenga_p3)
        	(gripper_jenga_1 ?gripper_extra_p1 ?jenga_p2)
        	(jenga_jenga_0 ?jenga_p1 ?jenga_p2)
        	(jenga_jenga_0 ?jenga_p3 ?jenga_p2)
        	(clear3_jenga_jenga_4 ?jenga_p1) 
        	(clear3_jenga_jenga_4 ?jenga_p2) 
        	(clear3_jenga_jenga_4 ?jenga_p3) 
        	(clear3_jenga_jenga_6 ?jenga_p1) 
        	(clear3_jenga_jenga_6 ?jenga_p2) 
        	(clear3_jenga_jenga_6 ?jenga_p3) 
    ) 
    :effect (and 
         	(loc_jenga_1 loc_jengaTarget_Const ?jenga_p2) 
        	(jenga_jenga_4 ?jenga_p1 ?jenga_p2) 
        	(jenga_jenga_6 ?jenga_p3 ?jenga_p2) 
        	(not (loc_jenga_0 loc_jengaTarget_Const ?jenga_p2))
        	(not (loc_jenga_2 loc_jengaTarget_Const ?jenga_p2))
        	(not (jenga_jenga_0 ?jenga_p1 ?jenga_p2))
        	(not (jenga_jenga_0 ?jenga_p3 ?jenga_p2))
        	(not (jenga_jenga_1 ?jenga_p1 ?jenga_p2))
        	(not (jenga_jenga_1 ?jenga_p3 ?jenga_p2))
        	(not (jenga_jenga_2 ?jenga_p1 ?jenga_p2))
        	(not (jenga_jenga_2 ?jenga_p3 ?jenga_p2))
        	(not (jenga_jenga_3 ?jenga_p1 ?jenga_p2))
        	(not (jenga_jenga_3 ?jenga_p3 ?jenga_p2))
        	(not (jenga_jenga_4 ?jenga_p3 ?jenga_p2))
        	(not (jenga_jenga_5 ?jenga_p1 ?jenga_p2))
        	(not (jenga_jenga_5 ?jenga_p3 ?jenga_p2))
        	(not (jenga_jenga_6 ?jenga_p1 ?jenga_p2))
        	(not (jenga_jenga_7 ?jenga_p1 ?jenga_p2))
        	(not (jenga_jenga_7 ?jenga_p3 ?jenga_p2))
        	(not (jenga_jenga_8 ?jenga_p1 ?jenga_p2))
        	(not (jenga_jenga_8 ?jenga_p3 ?jenga_p2))
        	(not (clear3_jenga_jenga_4 ?jenga_p1)) 
        	(not (clear3_jenga_jenga_6 ?jenga_p3)) 
    ) 
)

(:action a9 
    :parameters ( ?jenga_p2 - jenga  ?jenga_p3 - jenga  ?jenga_p1 - jenga  
            ?gripper_extra_p1 - gripper )
    :precondition (and 
        	(not (= ?jenga_p2 ?jenga_p3))
        	(not (= ?jenga_p2 ?jenga_p1))
        	(not (= ?jenga_p3 ?jenga_p1))
        	(loc_jenga_0 loc_jengaTarget_Const ?jenga_p1)
        	(loc_jenga_1 loc_jengaTarget_Const ?jenga_p2)
        	(loc_jenga_1 loc_jengaTarget_Const ?jenga_p3)
        	(gripper_jenga_0 ?gripper_extra_p1 ?jenga_p2)
        	(gripper_jenga_0 ?gripper_extra_p1 ?jenga_p3)
        	(gripper_jenga_1 ?gripper_extra_p1 ?jenga_p1)
        	(jenga_jenga_0 ?jenga_p2 ?jenga_p1)
        	(jenga_jenga_0 ?jenga_p3 ?jenga_p1)
        	(clear3_jenga_jenga_5 ?jenga_p1) 
        	(clear3_jenga_jenga_5 ?jenga_p2) 
        	(clear3_jenga_jenga_5 ?jenga_p3) 
        	(clear3_jenga_jenga_7 ?jenga_p1) 
        	(clear3_jenga_jenga_7 ?jenga_p2) 
        	(clear3_jenga_jenga_7 ?jenga_p3) 
    ) 
    :effect (and 
         	(loc_jenga_1 loc_jengaTarget_Const ?jenga_p1) 
        	(jenga_jenga_5 ?jenga_p2 ?jenga_p1) 
        	(jenga_jenga_7 ?jenga_p3 ?jenga_p1) 
        	(not (loc_jenga_0 loc_jengaTarget_Const ?jenga_p1))
        	(not (loc_jenga_2 loc_jengaTarget_Const ?jenga_p1))
        	(not (jenga_jenga_0 ?jenga_p2 ?jenga_p1))
        	(not (jenga_jenga_0 ?jenga_p3 ?jenga_p1))
        	(not (jenga_jenga_1 ?jenga_p2 ?jenga_p1))
        	(not (jenga_jenga_1 ?jenga_p3 ?jenga_p1))
        	(not (jenga_jenga_2 ?jenga_p2 ?jenga_p1))
        	(not (jenga_jenga_2 ?jenga_p3 ?jenga_p1))
        	(not (jenga_jenga_3 ?jenga_p2 ?jenga_p1))
        	(not (jenga_jenga_3 ?jenga_p3 ?jenga_p1))
        	(not (jenga_jenga_4 ?jenga_p2 ?jenga_p1))
        	(not (jenga_jenga_4 ?jenga_p3 ?jenga_p1))
        	(not (jenga_jenga_5 ?jenga_p3 ?jenga_p1))
        	(not (jenga_jenga_6 ?jenga_p2 ?jenga_p1))
        	(not (jenga_jenga_6 ?jenga_p3 ?jenga_p1))
        	(not (jenga_jenga_7 ?jenga_p2 ?jenga_p1))
        	(not (jenga_jenga_8 ?jenga_p2 ?jenga_p1))
        	(not (jenga_jenga_8 ?jenga_p3 ?jenga_p1))
        	(not (clear3_jenga_jenga_5 ?jenga_p2)) 
        	(not (clear3_jenga_jenga_7 ?jenga_p3)) 
    ) 
)

(:action a10 
    :parameters ( ?jenga_p1 - jenga  ?gripper_p1 - gripper )
    :precondition (and 
        	(gripper_jenga_2 ?gripper_p1 ?jenga_p1)
        	(clear3_gripper_jenga_1 ?gripper_p1) 
    ) 
    :effect (and 
         	(gripper_jenga_1 ?gripper_p1 ?jenga_p1) 
        	(not (gripper_jenga_0 ?gripper_p1 ?jenga_p1))
        	(not (gripper_jenga_2 ?gripper_p1 ?jenga_p1))
        	(clear3_gripper_jenga_2 ?gripper_p1) 
        	(not (clear3_gripper_jenga_1 ?gripper_p1)) 
    ) 
)

(:action a11 
    :parameters ( ?jenga_p1 - jenga  ?gripper_p1 - gripper )
    :precondition (and 
        	(gripper_jenga_0 ?gripper_p1 ?jenga_p1)
        	(clear3_gripper_jenga_1 ?gripper_p1) 
        	(clear3_gripper_jenga_2 ?gripper_p1) 
    ) 
    :effect (and 
         	(gripper_jenga_2 ?gripper_p1 ?jenga_p1) 
        	(not (gripper_jenga_0 ?gripper_p1 ?jenga_p1))
        	(not (gripper_jenga_1 ?gripper_p1 ?jenga_p1))
        	(not (clear3_gripper_jenga_2 ?gripper_p1)) 
    ) 
)

(:action a12 
    :parameters ( ?jenga_p1 - jenga  ?gripper_p1 - gripper )
    :precondition (and 
        	(loc_jenga_1 loc_jengaTarget_Const ?jenga_p1)
        	(gripper_jenga_1 ?gripper_p1 ?jenga_p1)
        	(clear3_gripper_jenga_2 ?gripper_p1) 
    ) 
    :effect (and 
         	(gripper_jenga_2 ?gripper_p1 ?jenga_p1) 
        	(not (gripper_jenga_0 ?gripper_p1 ?jenga_p1))
        	(not (gripper_jenga_1 ?gripper_p1 ?jenga_p1))
        	(clear3_gripper_jenga_1 ?gripper_p1) 
        	(not (clear3_gripper_jenga_2 ?gripper_p1)) 
    ) 
)

(:action a13 
    :parameters ( ?jenga_p1 - jenga  ?gripper_extra_p1 - gripper )
    :precondition (and 
        	(loc_jenga_0 loc_jengaTarget_Const ?jenga_p1)
        	(gripper_jenga_1 ?gripper_extra_p1 ?jenga_p1) 
    ) 
    :effect (and 
         	(loc_jenga_1 loc_jengaTarget_Const ?jenga_p1) 
        	(not (loc_jenga_0 loc_jengaTarget_Const ?jenga_p1))
    ) 
)

(:action a14 
    :parameters ( ?jenga_p1 - jenga  ?gripper_p1 - gripper )
    :precondition (and 
        	(loc_jenga_1 loc_jengaTarget_Const ?jenga_p1)
        	(gripper_jenga_2 ?gripper_p1 ?jenga_p1)
        	(clear3_gripper_jenga_1 ?gripper_p1) 
    ) 
    :effect (and 
         	(gripper_jenga_0 ?gripper_p1 ?jenga_p1) 
        	(not (gripper_jenga_1 ?gripper_p1 ?jenga_p1))
        	(not (gripper_jenga_2 ?gripper_p1 ?jenga_p1))
        	(clear3_gripper_jenga_2 ?gripper_p1) 
    ) 
))

\end{Verbatim}

\subsubsection{Packing a Box}

\begin{Verbatim}[breaklines=true]
(define (domain Packing)
(:requirements :strips :typing :equality :conditional-effects :existential-preconditions :universal-preconditions)
(:types 
	can
	gripper
	surface
)

(:predicates 
	(gripper_can_0 ?x - gripper ?y - can)
	(gripper_can_1 ?x - gripper ?y - can)
	(gripper_can_2 ?x - gripper ?y - can)
	(can_surface_0 ?x - can ?y - surface)
	(can_surface_1 ?x - can ?y - surface)
	(clear3_gripper_can_1 ?x - gripper)
	(clear3_gripper_can_2 ?x - gripper)
)

(:action a1 
:parameters ( ?can_p1 - can  ?gripper_p1 - gripper )
:precondition (and 
	(gripper_can_2 ?gripper_p1 ?can_p1)
	(clear3_gripper_can_1 ?gripper_p1) 
) 
:effect (and 
 	(gripper_can_1 ?gripper_p1 ?can_p1) 
	(not (gripper_can_0 ?gripper_p1 ?can_p1))
	(not (gripper_can_2 ?gripper_p1 ?can_p1))
	(clear3_gripper_can_2 ?gripper_p1) 
	(not (clear3_gripper_can_1 ?gripper_p1)) 
 ) 
)

(:action a2 
:parameters ( ?can_p1 - can  ?gripper_p1 - gripper )
:precondition (and 
	(gripper_can_0 ?gripper_p1 ?can_p1)
	(clear3_gripper_can_1 ?gripper_p1) 
	(clear3_gripper_can_2 ?gripper_p1) 
) 
:effect (and 
 	(gripper_can_2 ?gripper_p1 ?can_p1) 
	(not (gripper_can_0 ?gripper_p1 ?can_p1))
	(not (gripper_can_1 ?gripper_p1 ?can_p1))
	(not (clear3_gripper_can_2 ?gripper_p1)) 
 ) 
)

(:action a3 
:parameters ( ?can_p1 - can  ?surface_extra_p1 - surface  ?gripper_p1 - gripper )
:precondition (and 
	(can_surface_1 ?can_p1 ?surface_extra_p1)
	(gripper_can_2 ?gripper_p1 ?can_p1)
	(clear3_gripper_can_1 ?gripper_p1) 
) 
:effect (and 
 	(gripper_can_0 ?gripper_p1 ?can_p1) 
	(not (gripper_can_1 ?gripper_p1 ?can_p1))
	(not (gripper_can_2 ?gripper_p1 ?can_p1))
	(clear3_gripper_can_2 ?gripper_p1) 
 ) 
)

(:action a4 
:parameters ( ?can_p1 - can  ?surface_extra_p1 - surface  ?gripper_p1 - gripper )
:precondition (and 
	(can_surface_1 ?can_p1 ?surface_extra_p1)
	(gripper_can_1 ?gripper_p1 ?can_p1)
	(clear3_gripper_can_2 ?gripper_p1) 
) 
:effect (and 
 	(gripper_can_2 ?gripper_p1 ?can_p1) 
	(not (gripper_can_0 ?gripper_p1 ?can_p1))
	(not (gripper_can_1 ?gripper_p1 ?can_p1))
	(clear3_gripper_can_1 ?gripper_p1) 
	(not (clear3_gripper_can_2 ?gripper_p1)) 
 ) 
)

(:action a5 
:parameters ( ?can_p1 - can  ?gripper_extra_p1 - gripper  ?surface_p1 - surface )
:precondition (and 
	(can_surface_0 ?can_p1 ?surface_p1)
	(gripper_can_1 ?gripper_extra_p1 ?can_p1)
) 
:effect (and 
 	(can_surface_1 ?can_p1 ?surface_p1) 
	(not (can_surface_0 ?can_p1 ?surface_p1))
 ) 
))
\end{Verbatim}

\subsection{Code as Policies Evaluation Prompts}

\subsubsection{Delivering Items in a Cafe}

% \subsubsection*{Number of items: 1}
\subsubsection*{Domain-specific Prompt for Robot Actions}
\begin{promptbox}
\begin{Verbatim}[breaklines=true]
jointnames = (["torso_lift_joint", "shoulder_pan_joint", "shoulder_lift_joint", 
            "upperarm_roll_joint", "elbow_flex_joint", "forearm_roll_joint", 
            "wrist_flex_joint", "wrist_roll_joint"])

# define function: openGripper(robot)
def openGripper(robot):
    taskmanip = interfaces.TaskManipulation(robot)
    with robot:
        taskmanip.ReleaseFingers()
    robot.WaitForController(0)

# define function: grab_success_flag = grab_object(object_name)
def grab_object(object_to_grab):
    o = env.GetKinBody(object_to_grab)
    ot = o.GetTransform()
    robot_t = robot.GetLink("wrist_roll_link").GetTransform()
    euclidean_distance = np.linalg.norm(robot_t[:3,3]-ot[:3,3])
    obj_type = object_to_grab.split("_")[0]
    grab_range = [0.20,0.24]
    if euclidean_distance<grab_range[1] and euclidean_distance>grab_range[0]:
        robot.Grab(o)
        return True
    else:
        print("object out of grasp range")
        return False

# define function: un_grab(object_name)
def un_grab(object_name):
    robot.ReleaseAllGrabbed()

# define function: current_pose_of_object = get_current_pose_of_object(object_name)
def get_current_pose_of_object(object_name):
    obj = env.GetKinBody(object_name)
    obj_T = obj.GetTransform()
    obj_pose = get_pose_from_transform(obj_T)

    return obj_pose

# define function: robot_ik = get_ik(pose)
def get_ik(pose):
    end_effector_solution = get_transform_from_pose(pose)
    activate_manip_joints()
    # filter_option = IkFilterOptions.CheckEnvCollisions
    filter_option = IkFilterOptions.IgnoreEndEffectorCollisions
    
    with env:
        ikmodel = databases.inversekinematics.InverseKinematicsModel(robot,
                        iktype=IkParameterization.Type.Transform6D)

        if not ikmodel.load():
            ikmodel.autogenerate()
        try:
            solutions = ikmodel.manip.FindIKSolutions(end_effector_solution, filter_option)
        except:
            print("error")
            
    if len(solutions) == 0:
        print("NO IKs found, Probably Un-reachable transform")

    if len(solutions) > 0:
        if len(solutions) == 1:
            i = 0
        else:
            i = np.random.randint(0,len(solutions))
    else:
        return []
    
    return solutions[i]

# define function: go_to_gripper_pose(pose)
def go_to_gripper_pose(pose):
    activate_manip_joints()    
    robot.SetActiveDOFValues(pose)

# define function: go_to_base_pose(pose)
def go_to_base_pose(pose):
    activate_base_joints()
    robot.SetActiveDOFValues(pose)

# define function: current_arm_joint_values = get_current_arm_joint_values_of_robot()
def get_current_arm_joint_values_of_robot():
    activate_manip_joints()
    return robot.GetActiveDOFValues()

# define function: current_base_joint_values = get_current_base_joint_values()
def get_current_base_joint_values_of_robot():
    activate_base_joints
    return robot.GetActiveDOFValues()

# define function: pose = get_pose_from_transform(transform)
def get_pose_from_transform(transform):
    quat = poseFromMatrix(transform)[:4]
    eul = axisAngleFromQuat(quat)
    pose = []
    pose.extend(poseFromMatrix(transform)[4:])
    pose.extend(eul)

    return pose

# define function: collision_flag = collision_check(object_name)
def collision_check(object_name):
    collision_flag = False
    obj = env.GetKinBody(object_name)

    for obj2 in env.GetBodies():
        if obj != obj2:
            collision = env.CheckCollision(obj,obj2)
            if collision:
                if (obj2 == robot and obj in robot.GetGrabbed()) or 
                (obj == robot and obj2 in robot.GetGrabbed()):
                    collision = False
                    continue
                collision_flag = True
                break
    return collision_flag

# define function: activate_base_joints()
def activate_base_joints():
    robot.SetActiveDOFs([], DOFAffine.X | DOFAffine.Y | DOFAffine.RotationAxis)

# define function: activate_arm_joints()
def activate_manip_joints():
    robot.SetActiveDOFs([robot.GetJoint(name).GetDOFIndex() for name in jointnames])

# define function: grasp_pose_for_object = generate_grasp_pose(object_to_grab)
def generate_grasp_pose(object_to_grab):
    activate_manip_joints()
    rot_Z = matrixFromAxisAngle([0, 0, -np.pi/2])
    gripper_offset = -0.04
    world_T_obj = env.GetKinBody(object_to_grab).GetTransform()

    rot_ang = np.random.uniform(low = -np.pi, high = np.pi)
    obj_T_gripper = matrixFromPose([1, 0, 0, 0, gripper_offset, 0, 0.2/2.0])
    rot_mat = matrixFromAxisAngle([0, 0, rot_ang])

    wrist_roll_pose = robot.GetLink("wrist_roll_link").GetTransform()
    gripper_pose = robot.GetLink("gripper_link").GetTransform()
    wrist_pose_wrt_gripper = np.matmul(np.linalg.inv(gripper_pose), wrist_roll_pose)

    grasp_T = world_T_obj.dot(rot_mat).dot(rot_Z).dot(obj_T_gripper)
    grasp_T = np.matmul(grasp_T,wrist_pose_wrt_gripper)
    
    grasp_pose = get_pose_from_transform(grasp_T)

    return grasp_pose

# define function: transform = get_tranform_from_pose(pose)
def get_transform_from_pose(pose):
    quat = quatFromAxisAngle(pose[3:])
    pos = pose[:3]
    pose = []
    pose.extend(quat)
    pose.extend(pos)
    transform = matrixFromPose(pose)

    return transform


# define function: putdown_pose_for_object = 
            generate_put_down_pose(object_name,target_name)
def generate_put_down_pose(object_name,target_name):
    target_id = int(target_name.split("_")[1])
    obj = env.GetKinBody(object_name)
    can_radius = 0.03
    sampling_range = [[-0.175,0.175], [-0.175,0.175]]
    target_t = env.GetKinBody(target_name).GetTransform()

    x = np.random.uniform(low=sampling_range[0][0]+can_radius,
        high=sampling_range[0][1]-can_radius)
    y = np.random.uniform(low=sampling_range[1][0]+can_radius,
        high=sampling_range[1][1]-can_radius)


    
    target_id = int(target_name.split("_")[1])
    if target_id == 0:
        z = 0
        rot_z = np.eye(4)
    else:
        z = (0.1/2.0) + 0.005
        rot_angle = np.random.uniform(low=-np.pi,high=np.pi)
        rot_z = matrixFromAxisAngle([0,0,rot_angle])


    
    t = matrixFromPose([1,0,0,0,x,y,z])
    t = target_t.dot(t)
    t = t.dot(rot_z)

    rot_Z = matrixFromAxisAngle([0, 0, -np.pi/2])
    gripper_offset = -0.04
    world_T_obj = t

    rot_ang = np.random.uniform(low = -np.pi, high = np.pi)
    obj_T_gripper = matrixFromPose([1, 0, 0, 0, gripper_offset, 0, 0.2/2.0])
    rot_mat = matrixFromAxisAngle([0, 0, rot_ang])

    wrist_roll_pose = robot.GetLink("wrist_roll_link").GetTransform()
    gripper_pose = robot.GetLink("gripper_link").GetTransform()
    wrist_pose_wrt_gripper = np.matmul(np.linalg.inv(gripper_pose), wrist_roll_pose)

    grasp_T = world_T_obj.dot(rot_mat).dot(rot_Z).dot(obj_T_gripper)
    grasp_T = np.matmul(grasp_T,wrist_pose_wrt_gripper)
    
    pose = get_pose_from_transform(grasp_T)
   
    return pose

# define function: base_pose_around_surface = 
        generate_base_pose_around_surface(target_name)
def generate_base_pose_around_surface(target_name):
    target_id = int(target_name.split("_")[1])
    diff = 0.35
    x_dim = 0.45
    x_offset = -diff-x_dim
    y_offset = 0
    target_t = env.GetKinBody(target_name).GetTransform()
    diff_translation_matrix = matrixFromPose([1,0,0,0,x_offset,y_offset,0])

    if target_id == 0:
        rot_angle = (2*np.pi) / 4.0
    else:
        rot_angle = np.random.uniform(low=-np.pi,high=np.pi)
    
    rot_Z = matrixFromAxisAngle([0, 0, -np.pi/2])
    rot_mat = matrixFromAxisAngle([0,0,rot_angle])
    t = np.eye(4)
    t = target_t.dot(rot_mat).dot(rot_Z).dot(diff_translation_matrix)


    _x = t[0,3]
    _y = t[1,3]
    _yaw = axisAngleFromRotationMatrix(t[:3,:3])[-1]
    pose = [_x,_y,_yaw]

    return pose

# define function: tuck_arm()
def tuck_arm():
    activate_manip_joints()
    dof_values = [0, 1.32, 1.4, -0.2, 1.72, 0, 1.3599999999999999, 0.0]

    robot.SetActiveDOFValues(dof_values)

grabbed_armTuckDOFs = [0, 1.32, 1.4, -0.2, 1.72, 0, 1.3599999999999999, 0.0]

# example: tuck the robot arm to a tuck pose.
tuck_arm()


# example: go to "surface_1"
p = generate_base_pose_around_surface("surface_1")
go_to_base_pose(p)

# example: grab can_1 which is currently on "surface_0".
p = generate_base_pose_around_surface("surface_2")
go_to_base_pose(p)
p_g = generate_grasp_pose("can_1")
ik = get_ik(p_g)
if len(ik) != 0:
    go_to_gripper_pose(ik)
grabbed_flag = grab_object("can_1")

# example: pickup a grabbed object "can_2", which is on "surface_1" 
                when robot is near "surface_1".
p_g = generate_grasp_pose("can_2")
ik = get_ik(p_g)
if len(ik) != 0:
    go_to_gripper_pose(ik)
grabbed_flag = grab_object("can_2")
if grabbed_flag:
    tuck_arm()

# example: put down "can_1", which is already being held by the 
            robot gripper on "surface_1".
p = generate_base_pose_around_surface("surface_1")
go_to_base_pose(p)
p_g = generate_put_down_pose("can_1","surface_1")
ik = get_ik(p_g)
if len(ik) != 0:
    go_to_gripper_pose(ik)
un_grab("can_1")
tuck_arm()

# example: check collisions for "can_1".
collision_flag = collision_check("can_1")

# example: pickup can_1, which is on surface_0 and place it on surface_2
p = generate_base_pose_around_surface("surface_0")
go_to_base_pose(p)
p_g = generate_grasp_pose("can_1")
ik = get_ik(p_g)
if len(ik) != 0:
    go_to_gripper_pose(ik)
grabbed_flag = grab_object("can_1")
if grabbed_flag:
    tuck_arm()
p = generate_base_pose_around_surface("surface_2")
go_to_base_pose(p)
p_g = generate_put_down_pose("can_1","surface_2")
ik = get_ik(p_g)
if len(ik) != 0:
    go_to_gripper_pose(ik)
un_grab("can_1")
tuck_arm()


% \end{Verbatim}
\end{promptbox}

\subsubsection*{Prompt for Delivering 1 Item}
\begin{promptbox}

\begin{Verbatim}[breaklines=true]
Using the functions given above solve the following problem: #
There are 2 cans, can_1, and can_2, and three surfaces, surface_1, surface_2, 
and surface_0 in the environment. In the initial state, both the cans, can_1 and 
can_2 are on the surface - surface_2. the function returns True if robot has 
successfully picked up both cans from surface_2 and placed can_1 on surface_0 and 
can_2 on surface_1. success_flag is True if the function succeeds in task and 
there is no object in collision with any other object in the environemnt.
# The signature for the function required is:
success_flag = place_cans_on_goal_surfaces().
# only give the code and no other text including comments, markdowns or for which
language the code is given. Also, at the end call the function as well.
\end{Verbatim}
\end{promptbox}

\subsubsection*{Prompt for Delivering 2 Item}
\begin{promptbox}
\begin{Verbatim}[breaklines=true]
Using the functions given above solve the following problem: #
There are 2 cans, can_1, and can_2, and three surfaces, surface_1, surface_2,
and surface_0 in the environment. In the initial state, can_1 is on surface_2 
and can_2 is on surface_1. the function returns True if robot has successfully picked 
up can_1 from surface_2 and placed it on surface_1, and also has picked can_2 from 
surface_1 and placed it on surface_0. success_flag is True if the function succeeds
in task and there is no object in collision with any other object in the environemnt.
# The signature for the function required is:
success_flag = place_cans_on_goal_surfaces().
# only give the code and no other text including comments, markdowns or for which 
language the code is given. Also, at the end call the function as well.
\end{Verbatim}
\end{promptbox}

% \newpage

\subsubsection{Building Keva Structures}

\subsubsection*{Domain-specific Prompt for Robot Actions}

\begin{promptbox}
\begin{Verbatim}[breaklines=true]
# define function: success_flag = tuck_arm()
def tuck_arm(arm="left"):
    release(arm)
    if arm == "left":
        solution = left_arm_tuck_DOFs
    elif arm == "right":
        solution = right_arm_tuck_DOFs

    try:
        robot.SetActiveDOFValues(solution)
    except:
        pass
    
    openGripper()        
    return True

# define function: grab(object_name)
def grab(obj,arm="left"):
    o = env.GetKinBody(obj)
    if arm == "left":
        gripper_link = left_gripper_link
    elif arm == "right":
        gripper_link = right_gripper_link

    robot_t = gripper_link.GetTransform()
    ot = o.GetTransform()
    euclidean_distance = np.linalg.norm(robot_t[:3,3]-ot[:3,3])
    if euclidean_distance<grab_range[1] and euclidean_distance>grab_range[0]:
        robot.Grab(o)
    else:
        print("object out of grasp range")

# define function: release(object_name)
def release(object_name="object_name",arm="left"):
    robot.ReleaseAllGrabbed()

# define function: collision_flag = collision_check(object_name)
def collision_check(object_name):
    collision_flag = False
    obj = env.GetKinBody(object_name)

    for obj2 in env.GetBodies():
        if obj != obj2:
            collision = env.CheckCollision(obj,obj2)
            if collision:
                if (obj2 == robot and obj in robot.GetGrabbed()) or 
                (obj == robot and obj2 in robot.GetGrabbed()):
                    collision = False
                    continue
                collision_flag = True
                break
    
    return collision_flag

# define function: robot_ik = get_ik_solutions(pose)
def get_ik_solutions(end_effector_solution,robot_param="left"):
    end_effector_solution = get_transform_from_pose(end_effector_solution)
    current_state = robot.GetActiveDOFValues()
    collision = True
    if robot_param not in manipulator_groups:
        robot_param = "left"
    
    if robot_param == "left":
        ik_solver = left_ik_solver
    elif robot_param == "right":
        ik_solver = right_ik_solver

    ik_count = 0
    
    required_T = np.linalg.pinv(robot_init_transform).dot(end_effector_solution)
    pose = get_pose_from_transform(required_T)
    pos = pose[:3]
    orn = quatFromAxisAngle(pose[3:])

    while collision:
        seed_state = [np.random.uniform(-3.14, 3.14)] * ik_solver.number_of_joints
        joint_values = ik_solver.get_ik(seed_state,
                                                pos[0], pos[1], pos[2], 
                                                orn[1], orn[2], orn[3], orn[0] 
                                                )
        
        if joint_values is not None:
            robot.SetActiveDOFValues(joint_values)
            if collision_check(str(robot.GetName())):
                collision=True
            else:
                collision = False
        else:
            print("no joint_values")
            joint_values = []
            break
        
    robot.SetActiveDOFValues(current_state)
    return joint_values

# define function: pose = get_pose_from_transform(transform)
def get_pose_from_transform(transform):
    quat = poseFromMatrix(transform)[:4]
    eul = axisAngleFromQuat(quat)
    pose = []
    pose.extend(poseFromMatrix(transform)[4:])
    pose.extend(eul)

    return pose

# define function: go_to_gripper_pose(pose)
def go_to_gripper_pose(pose,arm="left"):
    robot.SetActiveDOFValues(pose)

# define function: current_pose_of_object = get_current_pose_of_object(object_name)
def get_current_pose_of_object(object_name):
    obj = env.GetKinBody(object_name)
    obj_T = obj.GetTransform()
    obj_pose = get_pose_from_transform(obj_T)

    return obj_pose

# define function: transform = get_tranform_from_pose(pose)
def get_transform_from_pose(pose):
    quat = quatFromAxisAngle(pose[3:])
    pos = pose[:3]
    pose = []
    pose.extend(quat)
    pose.extend(pos)
    transform = matrixFromPose(pose)

    return transform

# define function: p = sample_grasp_pose(object_name)
def sample_grasp_pose(object_name):
    world_T_obj = env.GetKinBody(object_name).GetTransform()
        
    world_T_robot = get_transform_from_pose(robot.GetActiveDOFValues())
    robot_T_world = np.linalg.inv(world_T_robot)

    obj_T_robot = np.eye(4)
    obj_T_robot[1,3]= grasping_offset
    
    t1 = orpy.matrixFromAxisAngle([ 0, -np.pi/2, 0])
    t2 = orpy.matrixFromAxisAngle([-np.pi/2, 0, 0])

    obj_T_robot = np.matmul(np.matmul(obj_T_robot,t1),t2)
    t = np.matmul(world_T_obj,obj_T_robot)
    pose = get_pose_from_transform(t)
    
    return pose

# define function: set_plank(object_name)
def set_plank(plank_name):
    plank = env.GetKinBody(plank_name)
    x_offsets=[0.15,0.45]
    y_offsets=[0.65,0.05]

    while True:
        t = np.eye(4)
        t[0,3] = np.random.uniform(low = -0.45+x_offsets[0], high = 0.45-x_offsets[1])
        t[1,3] = np.random.uniform(low = -0.45+y_offsets[0], high = 0.45-y_offsets[1])
        t[2,3] = 0.0135

        t1 = orpy.matrixFromAxisAngle([-np.pi/2, 0, 0])
        t = t.dot(t1)
        plank.SetTransform(t)
        t2 = orpy.matrixFromAxisAngle([0,-np.pi/2, 0])
        plank.SetTransform(t.dot(t2))
        if not(collision_check(plank_name)):
            break

# define function: p = get_goal_put_down_pose_for_plank_1()
def get_goal_put_down_pose_for_plank_1():
    world_T_obj = env.GetKinBody("goalLoc_1").GetTransform()
    
    obj_T_robot = np.eye(4)
    obj_T_robot[1,3]= grasping_offset
    
    t1 = orpy.matrixFromAxisAngle([ 0, -np.pi/2, 0])
    t2 = orpy.matrixFromAxisAngle([-np.pi/2, 0, 0])

    obj_T_robot = np.matmul(np.matmul(obj_T_robot,t1),t2)
    t = np.matmul(world_T_obj,obj_T_robot)
    pose = get_pose_from_transform(t)
    
    return pose

# define function: p = sample_plank_on_left_of_other_plank(plank_1,plank_2)
def sample_plank_on_left_of_other_plank(plank_1,plank_2):
    plank_1 = env.GetKinBody(plank_1)
    relative_transform = get_transform_from_pose([0, 0, 0.077474999073727457, 0, 0, 0])

    required_t = plank_1.GetTransform().dot(relative_transform)

    world_T_obj = required_t
        
    world_T_robot = get_transform_from_pose(robot.GetActiveDOFValues())
    robot_T_world = np.linalg.inv(world_T_robot)

    obj_T_robot = np.eye(4)
    obj_T_robot[1,3]= grasping_offset
    
    t1 = orpy.matrixFromAxisAngle([ 0, -np.pi/2, 0])
    t2 = orpy.matrixFromAxisAngle([-np.pi/2, 0, 0])

    obj_T_robot = np.matmul(np.matmul(obj_T_robot,t1),t2)
    t = np.matmul(world_T_obj,obj_T_robot)
    pose = get_pose_from_transform(t)
    
    return pose

# define function: p = sample_plank_on_right_of_other_plank(plank_1,plank_2)
def sample_plank_on_right_of_other_plank(plank_1,plank_2):
    plank_1 = env.GetKinBody(plank_1)
    relative_transform =
                get_transform_from_pose([0, 0, -0.077474999073727457, 0, 0, 0])

    required_t = plank_1.GetTransform().dot(relative_transform)

    world_T_obj = required_t
        
    world_T_robot = get_transform_from_pose(robot.GetActiveDOFValues())
    robot_T_world = np.linalg.inv(world_T_robot)

    obj_T_robot = np.eye(4)
    obj_T_robot[1,3]= grasping_offset
    
    t1 = orpy.matrixFromAxisAngle([ 0, -np.pi/2, 0])
    t2 = orpy.matrixFromAxisAngle([-np.pi/2, 0, 0])

    obj_T_robot = np.matmul(np.matmul(obj_T_robot,t1),t2)
    t = np.matmul(world_T_obj,obj_T_robot)
    pose = get_pose_from_transform(t)
    
    return pose

# define function: p = 
sample_plank_horizontally_on_top_of_other_two_vertical_planks
                ("plank_1","plank_2","plank_3")
def sample_plank_horizontally_on_top_of_other_two_vertical_planks
        (plank_1,plank_2,plank_3):
    plank_1 = env.GetKinBody(plank_1)
    relative_transform = get_transform_from_pose([0.059907747268674261, 0,
            0.04373009875416578, 0, 1.5707963267948959, 0])
    required_t = plank_1.GetTransform().dot(relative_transform)

    world_T_obj = required_t
        
    world_T_robot = get_transform_from_pose(robot.GetActiveDOFValues())
    robot_T_world = np.linalg.inv(world_T_robot)

    obj_T_robot = np.eye(4)
    obj_T_robot[1,3]= grasping_offset
    
    t1 = orpy.matrixFromAxisAngle([ 0, -np.pi/2, 0])
    t2 = orpy.matrixFromAxisAngle([-np.pi/2, 0, 0])

    obj_T_robot = np.matmul(np.matmul(obj_T_robot,t1),t2)
    t = np.matmul(world_T_obj,obj_T_robot)
    pose = get_pose_from_transform(t)
    
    return pose

# define function: p = 
sample_plank_horizontally_on_top_of_other_two_horizontal_planks_on_
                left_side("plank_1","plank_2","plank_3")
def sample_plank_horizontally_on_top_of_other_two_horizontal_planks_on_
                left_side(plank_1,plank_2,plank_3):
    plank_1 = env.GetKinBody(plank_1)
    relative_transform = get_transform_from_pose([0.04, -0.023439999999999989, 0.04, 
            0, 1.5707963267948963, 0])
    required_t = plank_1.GetTransform().dot(relative_transform)

    world_T_obj = required_t
        
    world_T_robot = get_transform_from_pose(robot.GetActiveDOFValues())
    robot_T_world = np.linalg.inv(world_T_robot)

    obj_T_robot = np.eye(4)
    obj_T_robot[1,3]= grasping_offset
    
    t1 = orpy.matrixFromAxisAngle([ 0, -np.pi/2, 0])
    t2 = orpy.matrixFromAxisAngle([-np.pi/2, 0, 0])

    obj_T_robot = np.matmul(np.matmul(obj_T_robot,t1),t2)
    t = np.matmul(world_T_obj,obj_T_robot)
    pose = get_pose_from_transform(t)
    return pose

# define function: p = 
sample_plank_horizontally_on_top_of_other_two_horizontal_planks_on_
            right_side("plank_1","plank_2","plank_3")
def sample_plank_horizontally_on_top_of_other_two_horizontal_planks_on_
            right_side(plank_1,plank_2,plank_3):
    plank_1 = env.GetKinBody(plank_1)
    relative_transform = get_transform_from_pose([-0.04, -0.023399999999999935, 
            0.04, 0, 1.5707963267948968, 0])
    
    required_t = plank_1.GetTransform().dot(relative_transform)

    world_T_obj = required_t
        
    world_T_robot = get_transform_from_pose(robot.GetActiveDOFValues())
    robot_T_world = np.linalg.inv(world_T_robot)

    obj_T_robot = np.eye(4)
    obj_T_robot[1,3]= grasping_offset
    
    t1 = orpy.matrixFromAxisAngle([ 0, -np.pi/2, 0])
    t2 = orpy.matrixFromAxisAngle([-np.pi/2, 0, 0])

    obj_T_robot = np.matmul(np.matmul(obj_T_robot,t1),t2)
    t = np.matmul(world_T_obj,obj_T_robot)
    pose = get_pose_from_transform(t)
    
    return pose

# define function: p = 
sample_plank_perpendicularly_on_top_of_a_horizontal_plank_on_its_left_side
    (plank_1,plank_2)
def sample_plank_perpendicularly_on_top_of_a_horizontal_plank_on_its_
            left_side(plank_1,plank_2):
    plank_1 = env.GetKinBody(plank_1)
    relative_transform = get_transform_from_pose([-0.033744900319561663, 
        -4.9568220254682954e-16, 0.066504776477813554, 1.0368663869534288e-14, 
            -1.5707963267948974, -9.9776971655314691e-15])
    
    required_t = plank_1.GetTransform().dot(relative_transform)

    world_T_obj = required_t
        
    world_T_robot = get_transform_from_pose(robot.GetActiveDOFValues())
    robot_T_world = np.linalg.inv(world_T_robot)

    obj_T_robot = np.eye(4)
    obj_T_robot[1,3]= grasping_offset
    
    t1 = orpy.matrixFromAxisAngle([ 0, -np.pi/2, 0])
    t2 = orpy.matrixFromAxisAngle([-np.pi/2, 0, 0])

    obj_T_robot = np.matmul(np.matmul(obj_T_robot,t1),t2)
    t = np.matmul(world_T_obj,obj_T_robot)
    pose = get_pose_from_transform(t)
    return pose
    
# define function: p = 
sample_plank_perpendicularly_on_top_of_a_horizontal_plank_on_its_
            right_side(plank_1,plank_2)
def sample_plank_perpendicularly_on_top_of_a_horizontal_plank_on_its_
            right_side(plank_1,plank_2):
    plank_1 = env.GetKinBody(plank_1)
    relative_transform = get_transform_from_pose([0.04373009875416739, 
        -4.7544147175681199e-16, 0.066334486007690402, 1.0368663869534288e-14,
        -1.5707963267948974, -9.9776971655314691e-15])
    
    required_t = plank_1.GetTransform().dot(relative_transform)

    world_T_obj = required_t
        
    world_T_robot = get_transform_from_pose(robot.GetActiveDOFValues())
    robot_T_world = np.linalg.inv(world_T_robot)

    obj_T_robot = np.eye(4)
    obj_T_robot[1,3]= grasping_offset
    
    t1 = orpy.matrixFromAxisAngle([ 0, -np.pi/2, 0])
    t2 = orpy.matrixFromAxisAngle([-np.pi/2, 0, 0])

    obj_T_robot = np.matmul(np.matmul(obj_T_robot,t1),t2)
    t = np.matmul(world_T_obj,obj_T_robot)
    pose = get_pose_from_transform(t)
    return pose

# example: place plank_1 and plank_2 parallel to each other
set_plank("plank_1")
p = sample_grasp_pose("plank_1")
ik = get_ik_solutions(p)
if len(ik) > 0:
    go_to_gripper_pose(ik)
grab("plank_1")
p = get_goal_put_down_pose_for_plank_1()
ik = get_ik_solutions(p)
if len(ik) > 0:
    go_to_gripper_pose(ik)
release()
set_plank("plank_2")
p = sample_grasp_pose("plank_2")
ik = get_ik_solutions(p)
if len(ik) > 0:
    go_to_gripper_pose(ik)
grab("plank_2")
p = sample_plank_on_left_of_other_plank("plank_1","plank_2")
ik = get_ik_solutions(p)
if len(ik) > 0:
    go_to_gripper_pose(ik)
release()

# example: place plank_3 horizontally on top of vertically placed plank_1 and plank_2
set_plank("plank_3")
p = sample_grasp_pose("plank_3")
ik = get_ik_solutions(p)
if len(ik) > 0:
    go_to_gripper_pose(ik)
grab("plank_3")
p = sample_plank_on_top_of_other_planks_in_pose_1("plank_1", "plank_2", "plank_3")
ik = get_ik_solutions(p)
if len(ik) > 0:
    go_to_gripper_pose(ik)
release()

# example: place plank_2 vertically on plank_1 on the left
set_plank("plank_2")
p = sample_grasp_pose("plank_2")
ik = get_ik_solutions(p)
if len(ik) > 0:
    go_to_gripper_pose(ik)
grab("plank_2")
p = sample_plank_perpendicularly_on_top_of_a_horizontal_plank_on_its_
        left_side("plank_1","plank_2")
ik = get_ik_solutions(p)
if len(ik) > 0:
    go_to_gripper_pose(ik)
release()

# example: place plank_2 vertically on plank_1 on the right
set_plank("plank_2")
p = sample_grasp_pose("plank_2")
ik = get_ik_solutions(p)
if len(ik) > 0:
    go_to_gripper_pose(ik)
grab("plank_2")
p = sample_plank_perpendicularly_on_top_of_a_horizontal_plank_on_its_
        right_side("plank_1","plank_2")
ik = get_ik_solutions(p)
if len(ik) > 0:
    go_to_gripper_pose(ik)
release()

# example: place plank_3 horizontally on top of horizontally placed 
        plank_1 and plank_2
set_plank("plank_3")
p = sample_grasp_pose("plank_3")
ik = get_ik_solutions(p)
if len(ik) > 0:
    go_to_gripper_pose(ik)
grab("plank_3")
p = sample_plank_on_top_of_other_planks_in_pose_2("plank_1", "plank_2", "plank_3")
ik = get_ik_solutions(p)
if len(ik) > 0:
    go_to_gripper_pose(ik)
release()

# example: place plank_3 horizontally on top of horizontally placed 
            plank_1 and plank_2
set_plank("plank_3")
p = sample_grasp_pose("plank_3")
ik = get_ik_solutions(p)
if len(ik) > 0:
    go_to_gripper_pose(ik)
grab("plank_3")
p = sample_plank_on_top_of_other_planks_in_pose_3("plank_1", "plank_2", "plank_3")
ik = get_ik_solutions(p)
if len(ik) > 0:
    go_to_gripper_pose(ik)
release()

# example: tuck the robot arm to a tuck pose.
tuck_arm()

# example: check collisions for plank_1
collision_flag = collision_check("plank_1")
\end{Verbatim}
\end{promptbox}

\subsubsection*{Prompt for Building Structure 1}
\begin{promptbox}
\begin{Verbatim}[breaklines=true]
Using the functions given above solve the following problem: #
There are 3 planks in the environment. The function should arrange these planks such 
that all the planks are kept at their corresponding goal locations. plank_1 being at 
goal location corresponding to itself. I.e., plank_1 and plank_2 are placed vertically 
on the table parallel from each other and plank_3 should be placed horizontally on 
both of these planks. The function returns true if the planks are placed in a 
collision-free configuration, otherwise returns false.
# The signature for the function required is: 
success_flag = place_planks_in_a_goal_structure().
# only give the code and no other text including comments, markdowns or for which 
language the code is given. Also, at the end call the function as well.
\end{Verbatim}
\end{promptbox}

\subsubsection*{Prompt for Building Structure 2}

\begin{promptbox}
\begin{Verbatim}[breaklines=true]
Using the functions given above solve the following problem: #
There are 4 planks in the environment. The function should arrange these planks 
such that all the planks are kept at their corresponding 
goal locations. plank_1 and  plank_2 are placed horizontally on top of the table, with 
plank_1 being at goal location corresponding to itself and plank_2 placed parallelly 
on left of plank_1. plank_4  is placed on the plank_2 and plank_1 horizontally on 
its side on the left and plank_3 is placed on the plank_2 and plank_1 horizontally on 
its side on the right. plank_3  is also placed parallelly on the left of the plank_4.
# The signature for the function required is:
success_flag = place_planks_in_a_goal_structure().
# only give the code and no other text including comments, markdowns or for which 
language the code is given. Also, at the end call the function as well.
\end{Verbatim}
\end{promptbox}

\subsubsection*{Prompt for Building Structure 3}

\begin{promptbox}
\begin{Verbatim}[breaklines=true]
Using the functions given above solve the following problem: #
There are 6 planks in the environment. The function should arrange these planks 
such that all the planks are kept at their corresponding goal locations such that when
kept together, the planks create a structure of a 2D drawing of a house.
# The signature for the function required is:
success_flag = place_planks_in_a_goal_structure().
# only give the code and no other text including comments, markdowns or for which 
language the code is given. Also, at the end call the function as well.
\end{Verbatim}
\end{promptbox}

% \newpage

\subsubsection*{Prompt for Building Structure 4}
\begin{promptbox}
\begin{Verbatim}[breaklines=true]
Using the functions given above solve the following problem: #
There are 6 planks in the environment. The function should arrange these planks 
such that all the planks are kept at their corresponding goal locations. 
plank_1 and plank_2 are placed horizontally on top of the table, with plank_1 being
at goal location corresponding to itself and plank_2 placed parallelly on left of 
plank_1. plank_3 is placed on the plank_2 and plank_1 horizontally on its side on 
the left and plank_4 is placed on the plank_2 and plank_1 horizontally on its 
side on the right. plank_4 is also placed parallelly on the left of the plank_3. 
plank_4 and plank_5 are kept parallelly and vertically on top of plank_3, with plank_4 
kept on right side and plank_5 placed on left side of plank_4. plank_6 is placed 
horizontally on plank_4 and plank_5.
# The signature for the function required is:
success_flag = place_planks_in_a_goal_structure().
# only give the code and no other text including comments, markdowns or for which 
language the code is given. Also, at the end call the function as well.
\end{Verbatim}
\end{promptbox}

\subsubsection{Packing a Box}
\subsubsection*{Domain-specific Prompt for Robot Actions}

\begin{promptbox}
\begin{Verbatim}[breaklines=true]

# define function: grab_object(object_to_grab)
def grab_object(object_to_grab):
    o = env.GetKinBody(object_to_grab)
    robot.Grab(o)

# define function: un_grab(object_name)
def un_grab(object_name):
    robot.ReleaseAllGrabbed()

# define function: grasp_pose_for_object = generate_grasp_pose(object_to_grab)
def generate_grasp_pose(object_to_grab):
    world_T_obj = env.GetKinBody(object_to_grab).GetTransform()
    obj_T_robot = np.eye(4)
    obj_T_robot[2,3] = 0.21
    t1 = matrixFromAxisAngle([0, 0, -np.pi/4.0])
    obj_T_robot = obj_T_robot.dot(t1)
    t = np.matmul(world_T_obj,obj_T_robot)
    pose = get_pose_from_transform(t)
    
    return pose

# define function: putdown_pose_for_object = generate_putdown_pose
                    (object_name,target_name)
def generate_putdown_pose(object_name,target_name):
    obj_dims = [0.0325,0.0325]
    droparea_dims = 0.075

    drop = env.GetKinBody(target_name)
    drop_t = drop.GetTransform()

    x_edge_offset = abs(droparea_dims-obj_dims[0])
    y_edge_offset = abs(droparea_dims-obj_dims[1])

    x = np.random.uniform(low=-x_edge_offset,high=x_edge_offset)
    y = np.random.uniform(low=-y_edge_offset,high=y_edge_offset)
    z = 0.001



    world_T_obj = matrixFromPose([1,0,0,0,x,y,z])
    world_T_obj = drop_t.dot(world_T_obj)

    obj_T_robot = np.eye(4)
    obj_T_robot[2,3] = 0.21
    t1 = matrixFromAxisAngle([0, 0, -np.pi/4.0])
    obj_T_robot = obj_T_robot.dot(t1)
    t = np.matmul(world_T_obj,obj_T_robot)
    pose = get_pose_from_transform(t)

    return pose

# define function: current_pose_of_object = get_current_pose_of_object(object_name)
def get_current_pose_of_object(object_name):
    obj = env.GetKinBody(object_name)
    obj_T = obj.GetTransform()
    obj_pose = get_pose_from_transform(obj_T)

    return obj_pose

# define function: robot_ik = get_ik(pose)
def get_ik(pose):
    return pose

# define function: go_to_pose(pose)
def go_to_pose(pose):
    ik = get_ik(pose)
    robot.SetActiveDOFValues(ik)

# define function: current_joint_values = get_current_joint_values_of_robot()
def get_current_joint_values_of_robot():
    return robot.GetActiveDOFValues()

# define function: pose = get_pose_from_transform(transform)
def get_pose_from_transform(transform):
    quat = poseFromMatrix(transform)[:4]
    eul = axisAngleFromQuat(quat)
    pose = []
    pose.extend(poseFromMatrix(transform)[4:])
    pose.extend(eul)

    return pose

# define function: collision_flag = collision_check(object_name)
def collision_check(object_name):
    collision_flag = False
    obj = env.GetKinBody(object_name)

    for obj2 in env.GetBodies():
        if obj != obj2:
            collision = env.CheckCollision(obj,obj2)
            if collision:
                if (obj2 == robot and obj in robot.GetGrabbed()) 
                    or (obj == robot and obj2 in robot.GetGrabbed()):
                    collision = False
                    continue
                collision_flag = True
                break
    
    return collision_flag

# example: go to grasp pose of can_1.
p = generate_grasp_pose("can_1")
go_to_pose(p)

# example: grab can_1.
grab_object("can_1")

# example: pickup can_1.
p = generate_grasp_pose("can_1")
go_to_pose(p)
grab_object("can_1")

# example: putdown can_1 in box.
p = generate_putdown_pose("can_1","box")
go_to_pose(p)
un_grab("can_1")

# example: check collisions for can_1.
collision_flag = collision_check("can_1")
\end{promptbox}
\subsubsection*{Prompt For Packing 1 Can}
\begin{promptbox}
Using the functions given above solve the following problem: #
There is 1 can, can_1, on the table. The function returns true if it can accommodate
all the cans in a box at a pose such that each can is not in collision with every 
other can in the box. The function needs to accommodate all objects in the box.
# The signature for the function required is: 
collision_flag = put_all_cans_in_box("box").
# only give the code and no other text including comments, markdowns or for which 
language the code is given. Also, at the end call the function as well.
\end{Verbatim}
\end{promptbox}

\subsubsection*{Prompt For Packing 2 Can}
\begin{promptbox}
\begin{Verbatim}[breaklines=true]
Using the functions given above solve the following problem: #
There are 2 cans, can_1, and can_2, on the table. The function returns true if it can 
accommodate all the cans in a box at a pose such that each can is not in collision 
with every other can in the box. The function needs to accommodate all objects in 
the box.
# The signature for the function required is:
collision_flag = put_all_cans_in_box("box").
# only give the code and no other text including comments, markdowns or for which 
language the code is given. Also, at the end call the function as well.
\end{Verbatim}
\end{promptbox}

\subsubsection*{Prompt For Packing 3 Can}
\begin{promptbox}
\begin{Verbatim}[breaklines=true]
Using the functions given above solve the following problem: #
There are 3 cans, can_1, can_2, and can_3 on the table. The function returns true 
if it can accommodate all the cans in a box at a pose such that each can is not
in collision with every other can in the box. The function needs to accommodate all
objects in the box.
# The signature for the function required is:
collision_flag = put_all_cans_in_box("box").
# only give the code and no other text including comments, markdowns or for which 
language the code is given. Also, at the end call the function as well.
\end{Verbatim}
\end{promptbox}

\subsubsection*{Prompt For Packing 4 Can}
\begin{promptbox}
\begin{Verbatim}[breaklines=true]
Using the functions given above solve the following problem: #
There are 3 cans, can_1, can_2, and can_3 on the table. The function returns true
if it can accommodate all the cans in a box at a pose such that each can is not 
in collision with every other can in the box. The function needs to accommodate all
objects in the box.
# The signature for the function required is:
collision_flag = put_all_cans_in_box("box").
# only give the code and no other text including comments, markdowns or for which 
language the code is given. Also, at the end call the function as well.
\end{Verbatim}
\end{promptbox}

% \bibliographystyle{plainnat}
% \bibliography{ref}

\end{document}